\documentclass[nohyperref]{article}

\usepackage{microtype}
\usepackage{graphicx}
\usepackage{subfigure}
\usepackage{booktabs} 

\usepackage{hyperref}

\usepackage[ruled,linesnumbered,boxed]{algorithm2e}

\usepackage[accepted]{icml2023}

\usepackage{amsmath}
\usepackage{amssymb}
\usepackage{mathtools}
\usepackage{amsthm}

\usepackage{algorithm}
\usepackage{algorithmic}

\usepackage{booktabs} 

\usepackage[english]{babel}
\usepackage{moresize}
\usepackage{amsmath}

\usepackage{balance}
\usepackage{comment}
\usepackage{paralist}
\usepackage{bm}
\usepackage{pgfplots}
\usetikzlibrary{pgfplots.dateplot}

\usepackage{flushend}
\usepackage[english]{babel}
\usepackage[latin1]{inputenc}
\usepackage{mathrsfs}
\usepackage{graphicx}

\usepackage{amssymb}
\usepackage{amsfonts}
\usepackage{url}
\usepackage{longtable}
\usepackage{rotating}
\usepackage{multirow}
\usepackage{mathrsfs}
\usepackage{subfigure}
\usepackage{enumitem}
\usepackage{adjustbox}
\usepackage{hyperref}
\usepackage{bbding}
\usepackage{threeparttable}

\graphicspath{{imag/}}

\usepackage{pgfplots}
\usetikzlibrary{pgfplots.dateplot}

\usepackage{filecontents}
\definecolor{tblue}{RGB}{31,119,180}
\definecolor{torange}{RGB}{255,127,14}
\definecolor{tgreen}{RGB}{44,160,44}
\definecolor{tred}{RGB}{214,39,40}
\definecolor{tpurple}{RGB}{148,103,189}

\newcommand{\hide}[1]{} 

\newcommand{\ie}{\textit{i}.\textit{e}.}
\newcommand{\eg}{\textit{e}.\textit{g}.} 
\newcommand{\wrt}{\textit{w}.\textit{r}.\textit{t}}

\usepackage[capitalize,noabbrev]{cleveref}

\def\model{GraphST}

\theoremstyle{plain}

\theoremstyle{definition}

\theoremstyle{remark}

\usepackage[textsize=tiny]{todonotes}

\icmltitlerunning{Spatial-Temporal Graph Learning with Adversarial Contrastive Adaptation}
\begin{document}

\twocolumn[
\icmltitle{Spatial-Temporal Graph Learning with Adversarial Contrastive Adaptation}




\begin{icmlauthorlist}
\icmlauthor{Qianru Zhang}{yyy}
\icmlauthor{Chao Huang*}{yyy}
\icmlauthor{Lianghao Xia}{yyy}
\icmlauthor{Zheng Wang}{xxx}
\icmlauthor{Siuming Yiu}{yyy}
\icmlauthor{Ruihua Han}{yyy}
\end{icmlauthorlist}

\icmlaffiliation{yyy}{The Univeristy of Hong Kong, Hong Kong}
\icmlaffiliation{xxx}{Nanyang Technological University, Singapore}

\icmlcorrespondingauthor{Chao Huang}{chaohuang75@gmail.com}

\icmlkeywords{Machine Learning, ICML}

\vskip 0.3in
]



\printAffiliationsAndNotice{}  

\begin{abstract}
Spatial-temporal graph learning has emerged as a promising solution for modeling structured spatial-temporal data and learning region representations for various urban sensing tasks such as crime forecasting and traffic flow prediction. However, most existing models are vulnerable to the quality of the generated region graph due to the inaccurate graph-structured information aggregation schema. The ubiquitous spatial-temporal data noise and incompleteness in real-life scenarios pose challenges in generating high-quality region representations. To address this challenge, we propose a new spatial-temporal graph learning model (\model) for enabling effective self-supervised learning. Our proposed model is an adversarial contrastive learning paradigm that automates the distillation of crucial multi-view self-supervised information for robust spatial-temporal graph augmentation. We empower \model\ to adaptively identify hard samples for better self-supervision, enhancing the representation discrimination ability and robustness. In addition, we introduce a cross-view contrastive learning paradigm to model the inter-dependencies across view-specific region representations and preserve underlying relation heterogeneity. We demonstrate the superiority of our proposed \model\ method in various spatial-temporal prediction tasks on real-life datasets. We release our model implementation via the link: \url{https://github.com/HKUDS/GraphST}.

\end{abstract}


\vspace{-0.1in}
\section{Introduction}
\label{sec:intro}


Spatial-temporal graph representation learning aims to provide a meaningful latent embedding for each region in a city, which facilitates various spatial-temporal prediction tasks such as crime forecasting for public security~\cite{li2022spatial}, traffic flow prediction in intelligent transportation systems~\cite{pan2019urban,zheng2020gman}, and Point-of-Interest (POI) recommendation in location-based services~\cite{zhou2019topic}. One of the most promising solutions to this problem is the adoption of Graph Neural Networks (GNNs)~\cite{zhang2021multi,wu2022multi_graph}. These GNN approaches model region correlations with message passing over region graphs generated based on collected spatial-temporal data, such as human mobility traces and citywide traffic flow.


Although spatial-temporal graph neural models are highly effective, their reliance on the quality of the constructed region graph presents significant challenges. The current schema~\cite{zhang2020spatio,jin2020graph,zhou2023greto} of embedding propagation and refinement along graph structural connections is limited and often unable to capture the complexities of real-world urban environments. Data noise and incompleteness are also common in spatial-temporal data analysis, which further undermines the quality of the constructed region graph. For example, sensor readings may be lost or inaccurate~\cite{yi2016st}, and human mobility data from crowd sensors is frequently noisy~\cite{feng2019dplink}. In addition, spatially adjacent regions may not be strongly correlated due to differences in urban function, while similar regions may be far apart geographically~\cite{zhou2020riskoracle}. These factors contribute to the questionable quality of region-wise connection graphs, making it difficult to effectively perform spatial-temporal graph representation learning in the presence of noise perturbation.

With the recent success of self-supervised learning in mitigating data scarcity and noise challenges, we draw inspiration to propose a spatial-temporap graph pre-training framework with effective data augmentation, by exploring the following questions in model design.\vspace{-0.15in}
\begin{itemize}[leftmargin=*]
\item \textbf{Q1}: How can we offer self-supervised signals as spatial-temporal graph pre-training tasks?\vspace{-0.05in}
\item \textbf{Q2}: How can we automatically identify hard samples during contrastive learning to enhance model robustness?\vspace{-0.05in}
\item \textbf{Q3}: How can we model the inter-dependencies across different region relation views? \vspace{-0.05in}
\end{itemize}


To address these challenges, we propose a novel model called \model, which advances spatial-temporal graph representation learning by distilling self-supervisory information for pre-training. To generate self-supervisory signals over our multi-view region graph in an adaptive manner, we design a learnable function that reflects the global urban context across different regions in a city. Furthermore, we develop an adversarial contrastive learning model that endows our \model\ with the ability to automatically identify hard positive and negative samples under a contrastive min-max optimization framework. Lastly, we introduce a cross-view contrastive learning method that captures inter-view dependencies and enhances representation uniformity of different regions through a self-discrimination scheme.

Our work makes the following contributions:\vspace{-0.15in}
\begin{itemize}[leftmargin=*]

\item We investigate the drawbacks of existing spatial-temporal graph learning methods with non-robustness against the perturbation of noisy and incomplete urban data.\vspace{-0.05in}

\item We propose a spatial-temporal graph pre-training model \model, in which an adversarial contrastive paradigm is designed to augment region relationship learning with graph structure-adaptive self-supervision.\vspace{-0.05in}

\item We demonstrate the significant improvements that \model\ achieves over state-of-the-art baselines in different settings and spatial-temporal prediction tasks. \vspace{-0.05in}


\end{itemize}



\section{Preliminaries}
\label{sec:pre}

Our approach begins by partitioning the geographical area of a city into $J$ spatial regions, indexed by $j$ (e.g., $r_j$). To incorporate diverse urban contextual information into our latent embedding space, we propose a region embedding paradigm that draws on different heterogeneous data sources in urban space. Specifically, our model input is elaborated:

\noindent \textbf{Region Point-of-Interests (POIs)}. To characterize the region urban functions (\eg, accommodation, entertainment, education, medical and health), we construct a POI matrix $\mathcal{P} \in \mathbb{R}^{J\times C}$, in which $C$ is the number of POI categories.

\noindent \textbf{Human Mobility Trajectories}. To capture the dynamic urban flow among citywide regions in $T$ time slots, we generate a human trajectory set denoted as $\mathcal{M}$. This set contains individual human mobility traces $(r_s, r_d, t_s, t_d)$, which record the source and destination region $r_s$ and $r_d$ along with the timestamp information $t_s$ and $t_d$, respectively. 

\noindent \textbf{Problem Statement}. Using the input of the POI matrix $\mathcal{P}$, human trajectory set $\mathcal{M}$, and spatial distance matrix $\mathcal{D}$, the output of this task is the low-dimensional embeddings of all regions in a city. These embeddings represent the encoded representation of each region and can be used for downstream spatial-temporal mining tasks, such as traffic prediction and urban crime forecasting.

\section{Methodology}
\label{sec:solution}

\begin{figure*}
    \centering
\includegraphics[width=0.99\textwidth]{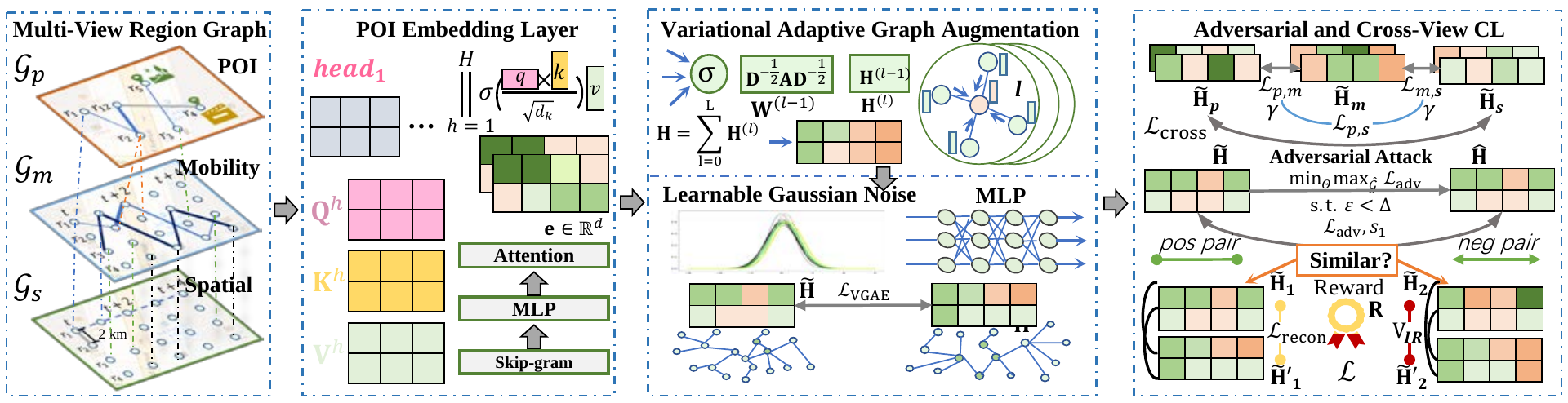}
    \vspace{-0.12in}
    \caption{Illustration of \model's overall architecture. (i) POI embedding layer is a parameterized function which maps spatial functional semantics into latent embedding space. ii) Variational adaptive graph augmentation is performed over multi-view spatial-temporal graph $\mathcal{G}$ with learnable structure corruption. iii) Adversarial contrastive augmentation generates self-supervision signals with hard samples for enhancing model optimization. The quality of training samples with contrastive augmentation will be improved in spatial-temporal graph space. iv) Cross-view contrastive learning aims to preserve the implicit region inter-dependencies from heterogeneous relational views.}
    \vspace{-0.1in}
    \label{fig:fram}
\end{figure*}

We introduce our spatial-temporal graph learning method \model\ with the key components illustrated in Figure~\ref{fig:fram}. 

\subsection{Graph-enhanced Spatial-Temporal Learning}
We develop a graph-enhanced spatial-temporal learning framework that is specifically designed to extract the underlying spatial-temporal relationships between regions by comprehensively capturing dynamic region dependencies across space and time from different perspectives such as POI semantics, urban flow transitions, and geo-locations. \vspace{-0.05in}

Our approach to preserving POI semantics in the embedding space involves using Skip-gram and MLP to embed the POI matrix $\mathcal{P} \in \mathbb{R}^{J\times C}$ into a set of latent representations. Specifically, we apply Skip-gram to capture the context of each POI and then use an MLP to transform these contextual embeddings into the final POI embeddings. This approach allows us to capture the semantic relatedness between POIs and maintain this information in the embedding space. In the resulting matrix $\bar{\textbf{E}} \in \mathbb{R}^{J\times d}$, each row corresponds to the $d$-dimensional vector of an individual region $r_j$.


\noindent\textbf{Multi-View Region Graph}. Prior to integrating multi-view information in a unified framework, we initially process each view independently and create view-specific graphs:\vspace{-0.05in}

\noindent i) \emph{The POI-based region graph $\mathcal{G}_p$} is generated by measuring the similarities between region POI semantic embeddings $\textbf{e}_j$ and $\textbf{e}_{j'}$. In this graph, two regions are connected if their cosine similarity $\cos(\textbf{e}_j, \textbf{e}_{j'})>\varepsilon$ exceeds a threshold $\varepsilon$. ii) \emph{The time-aware mobility-based graph $\mathcal{G}_m$} is constructed with nodes and edges across both space and time. For each region $r_j$, a sequence of time-aware region nodes is generated corresponding to different time slots ($t\in T$). The source region node $r_s^{t_s}$ at time slot $t_s$ is connected to the destination region node $r_d^{t_d}$ at time slot $t_d$ given the trajectory $(r_s, r_d, t_s, t_d)$.  iii) \emph{Geographic-based graph $\mathcal{G}_s$} connects spatially adjacent regions based on their distance.

We generate a unified multi-view region graph $\mathcal{G}$ by stacking the view-specific region graphs, namely $\mathcal{G}_p$, $\mathcal{G}_m$, $\mathcal{G}_s$, to reflect the region-wise relation heterogeneity. To achieve this, we add self-connection edges between nodes of the same region in different view-specific graph structures.\\\vspace{-0.12in}

\noindent \textbf{Information Propagation Paradigm}.
To capture both intra-region and inter-region relations in multi-view data, \model\ utilizes message passing over the multi-view region graph $\mathcal{G}$ in both time and space dimensions. The embedding propagation from the $(l-1)$-th graph layer to the $(l)$-th layer is defined as follows:
\begin{align}
    \textbf{h}_j^{(l)} = \sigma(\sum_{j'\in\mathcal{N}_j} \beta_{j,j'} \textbf{W}^{(l-1)} \textbf{h}_{j'}^{(l-1)})
\end{align}
\noindent $\textbf{h}_j$ represents the refined embedding vector of region $r_j$. The coefficient $\beta_{j,j'}$ is computed as $1/\sqrt{|\mathcal{N}_j||\mathcal{N}_{j'}|}$, where $\mathcal{N}_j$ denotes the set of neighboring nodes of region $r_j$ in graph $\mathcal{G}$. The initial embedding vector $\textbf{h}_j^{(0)}$ is derived from the POI embedding layer. The ReLU activation function is denoted by $\sigma(\cdot)$, and $\textbf{W}^{(l-1)}\in\mathbb{R}^{d\times d}$ represents the learnable transformation weights for the $(l-1)$-th iteration. Information aggregation is performed as:
\begin{align}
    \label{eq:multorder}
    \textbf{H} = \sum_{l=0}^L \textbf{H}^{(l)};
    \textbf{H}^{(l)} = \sigma(\textbf{D}^{-\frac{1}{2}} \textbf{A} \textbf{D}^{-\frac{1}{2}} \textbf{H}^{(l-1)}\textbf{W}^{(l-1)\top})
\end{align}
\noindent $\textbf{H}\in\mathbb{R}^{|\mathcal{V}|\times d}$ is the embedding matrix of all nodes in graph $\mathcal{G}$, where $|\mathcal{V}|$ is the number of nodes. The variable $L$ is the depth of the graph neural layers. The matrices $\textbf{A}\in\mathbb{R}^{|\mathcal{V}|\times \mathcal{V}|}$ and $\textbf{D}\in\mathbb{R}^{|\mathcal{V}|\times \mathcal{V}|}$ is the adjacent matrix of $\mathcal{G}$ with self-connections, and the diagonal degree matrix, respectively.

\subsection{Graph Augmentation}
Many existing graph contrastive learning methods use handcrafted contrastive views, such as stochastic graph structure corruption and masking, to simplify the model design. To make graph contrastive learning adaptable to global spatial-temporal relation heterogeneity, we propose an adaptive graph augmentation scheme for the multi-view region graph $\mathcal{G}$. Our scheme is based on the variational graph auto-encoder (VGAE) for self-supervision. This approach allows us to improve the performance of spatial-temporal representations by unifying contrastive and generative SSL to incorporate global spatial-temporal relation heterogeneity.


We propose a method to automatically learn the dependencies among different regions with global context and improve the region graph structures against noise perturbations such as low correlations between adjacent regions and strong dependencies between long-range regions. Our approach is based on the embedding mapping function $\mathcal{G}\rightarrow\mathbb{R}^{|\mathcal{V}|\times d}$ (Eq~\ref{eq:multorder}) and we perform Gaussian noise-based augmentation~\cite{rusak2020simple} with the following design:
\begin{align}
    \label{eq:lowdimGen}
    \tilde{\textbf{H}} = \mathbf{\Gamma} \cdot \text{MLP}_\text{std}(\textbf{H}) + \text{MLP}_\text{mean}(\textbf{H});\tilde{\mathcal{G}} = \eta(\tilde{\textbf{H}})
\end{align}
\noindent The augmented region embeddings for graph structure learning over all regions are denoted by $\tilde{\textbf{H}}\in\mathbb{R}^{|\mathcal{V}|\times d}$. The structure learning function, $\eta(\cdot)$, estimates the region-wise dependencies. The noise matrix, $\mathbf{\Gamma}\in\mathbb{R}^{|\mathcal{V}|\times d}$, is represented by elements $\gamma$ that are drawn from a Gaussian distribution with mean value $\mu$ and standard deviation $\sigma$.

To enable learnable contrastive view generation, \model\ uses two-layer MLPs with trainable parameters (\ie~$\text{MLP}_\text{mean}(\cdot)$ and $\text{MLP}_\text{std}(\cdot)$) to calculate the mean and standard deviation from the original embedding matrix $\textbf{H}$. Following this, the region graph structure learning function $\eta(\cdot)$ is repeatedly applied, and the adaptive graph augmentation is performed using a contrastive loss.
\begin{align}
    \mathcal{L}_\text{VGAE} = \sum\nolimits_j -\log\frac{\exp (\cos(\tilde{\textbf{h}}_j, \tilde{\textbf{h}}_j')/\tau)}{\sum\nolimits_{j'} \exp(\cos(\tilde{\textbf{h}}_j, \tilde{\textbf{h}}_{j'}')/\tau)}
\end{align}
\noindent $\tau$ is the temperature parameter to control the gradient effect.

\subsection{Adversarial Contrasting with Hard Samples}
\label{subsec:hard}
To enhance the robustness of our graph augmentation against perturbations, we incorporate adversarial self-supervision to identify hard negative and positive node samples. This improves our contrastive learning paradigm by providing auxiliary self-supervision signals that are beneficial for gradient learning during model training. By performing adversarial contrastive learning, we aim to improve the performance of our model and make it more robust to perturbations.

To achieve this goal, we perform adversarial data augmentation by maximizing the contrastive loss between the generative autoencoder view $\tilde{\mathcal{G}}$ and a constrained adversarial view $\hat{\mathcal{G}}$. The adversarial contrastive learning is performed under the mini-max adversarial optimization paradigm as follows: 
\begin{align}
\min\nolimits_{\mathbf{\Theta}} \mathop{\max}\nolimits_{\hat{\mathcal{G}}}~ \mathcal{L}_\text{adv} (\tilde{\mathcal{G}}, \hat{\mathcal{G}}),~~~
\text{s.t.}~ \varepsilon(\tilde{\mathcal{G}}, \hat{\mathcal{G}}) < \triangle
\end{align}
\noindent In the above equation, $\mathcal{L}_\text{adv}$ is the InfoNCE-based contrastive loss. $\mathbf{\Theta}$ represents the trainable model parameters of our variational graph auto-encoder. The error function $\varepsilon(\cdot)$ is defined as the absolute error between the adjacent matrices and the node features. $\triangle$ denotes hyperparameters for the error with respect to the adjacent matrix $\hat{\textbf{A}}, \tilde{\textbf{A}}$ and node embeddings $\hat{\textbf{H}}, \tilde{\textbf{H}}$.

To enable adversarial augmentation over the encoded node embeddings and graph structures, we adopt the projected gradient descent (PGD) attack~\cite{feng2022adversarial} for data perturbation. We define a supplement matrix for the adjacency matrix, denoted as $\hat{\mathbf{A}}''$, and $\hat{\mathbf{A}}' = \mathbf{1}_{|\hat{\mathcal{V}}'| \times |\hat{\mathcal{V}}'|} - \mathbf{I}_{|\hat{\mathcal{V}}'|}- \hat{\mathbf{A}}''$, where $\mathbf{1}_{|\hat{\mathcal{V}}'| \times |\hat{\mathcal{V}}'|}$ is an all-one matrix with the size of ${|\hat{\mathcal{V}}'| \times |\hat{\mathcal{V}}'|}$. After PGD attack, the perturbed adjacency matrix is given by $\hat{\mathbf{A}} = \hat{\mathbf{A}}' + (\bar{\mathbf{A}}-\hat{\mathbf{A}}') \circ \mathbf{L}_{\hat{\mathbf{A}}'}$, where $\circ$ denotes element-wise product. Each element $\mathbf{L}_{\hat{\mathbf{A}}'}[i,j]$ in $\mathbf{L}_{\hat{\mathbf{A}}'} \in \left\{ 0,1\right\}_{|\mathcal{V}| \times |\mathcal{V}|}$ represents the corresponding modification of the edge between node $v_i$ and node $v_j$. The perturbation on $\hat{\mathbf{H}}'$ is presented as $\hat{\mathbf{H}} = \hat{\mathbf{H}}' + \mathbf{L}_{\hat{\mathbf{H}}'}$. where $\mathbf{L}_{\hat{\mathbf{H}}} \in \mathbb{R}^{|\mathcal{V}|\times d}$ is the perturbation on feature matrix. For optimization relaxation, the convex hull $\ddot{\mathbf{L}}_{\hat{\mathbf{A}}'}$ replaces $\mathbf{L}_{\hat{\mathbf{A}}'}$, subject to the following constraints:

\begin{align}
    \label{eq:adv_constrains}
    &\mathcal{K}_{\hat{\mathbf{A}}'} = \left\{\ddot{\mathbf{L}}_{\hat{\mathbf{A}}'}| \sum\limits_{i,j}\ddot{\mathbf{L}}_{\hat{\mathbf{A}}'} \leq \triangle_{\hat{\mathbf{A}}'}, \ddot{\mathbf{L}}_{\hat{\mathbf{A}}'} \in [0,1]^{|\hat{\mathcal{V}}'| \times |\hat{\mathcal{V}}'|}\right\} \nonumber\\
    &\mathcal{K}_{\hat{\mathbf{H}}'} = \left\{\mathbf{L}_{\hat{\mathbf{H}}'}| \Vert \mathbf{L}_{\hat{\mathbf{H}}'} \Vert_{\infty} \leq \delta_{\hat{\mathbf{H}}'}, \mathbf{L}_{\hat{\mathbf{H}}'} \in \mathbb{R}^{|\hat{\mathcal{V}}'| \times d}\right\}
\end{align}
where $\delta_{\hat{\mathbf{H}}'}$ denotes the constraint on feature perturbation. During each iteration, the update is presented as:
\begin{align}
    &\ddot{\mathbf{L}}^{e}_{\hat{\mathbf{A}}'} = \prod \limits_{\mathbf{K}_{\hat{\mathbf{A}}'}}[\ddot{\mathbf{L}}^{e-1}_{\hat{\mathbf{A}}'} + \zeta \mathbf{G}_{\hat{\mathbf{A}}'}^{e}] \nonumber;\mathbf{G}_{\hat{\mathbf{A}}'}^{e} = \bigtriangledown_{\ddot{\mathbf{L}}^{e-1}_{\hat{\mathbf{A}}'}}\mathcal{L}'\nonumber\\
    & \mathbf{L}^{e}_{\hat{\mathbf{H}}'} =  \prod \limits_{\mathbf{K}_{\hat{\mathbf{H}}'}}[\mathbf{L}^{e-1}_{\hat{\mathbf{H}}'} + \eta \text{sgn}(\mathbf{G}_{\hat{\mathbf{H}}'}^{e})];\mathbf{H}_{\mathbf{X}}^{e} = \bigtriangledown_{\mathbf{L}^{e-1}_{\hat{\mathbf{H}}'}}\mathcal{L}' \nonumber
\end{align}
where $\mathbf{G}_{\hat{\mathbf{A}}'}^{e}$ and $\mathbf{G}_{\hat{\mathbf{H}}'}^{e}$ denotes the gradient of loss in terms of $\ddot{\mathbf{L}}^{e-1}_{\hat{\mathbf{A}}'}$ and $\mathbf{L}^{e-1}_{\hat{\mathbf{H}}'}$ on $e$-th iteration. 
  In addition, $\mathcal{L}'$ represents $\mathcal{L}_{\text{adv}}(\tilde{\mathbf{H}}, \hat{\mathbf{H}}^{e-1})$. $\prod \limits_{\mathbf{K}_{\hat{\mathbf{H}}'}}$ denotes the projection head and maps $\mathbf{L}_{\hat{\mathbf{H}}'}$ into $[-\delta_{\hat{\mathbf{H}}'},\delta_{\hat{\mathbf{H}}'}]$. Besides, Let $z = \sum\limits_{i,j}P_{[0,1]}[\mathbf{Y}-\vartheta \mathbf{1}_{{|\hat{\mathcal{V}}'|} \times |\hat{\mathcal{V}}'|}]$ and $\prod \limits_{\mathcal{K}_{\hat{\mathbf{A}}'}}(\mathbf{Y})$ is defined:
\begin{align}
\prod \limits_{\mathcal{K}_{\hat{\mathbf{A}}'}}(\mathbf{Y}) = \left\{
    \begin{aligned} 
    &P_{[0,1]}[\mathbf{Y}-\vartheta \mathbf{1}_{{|\hat{\mathcal{V}}'|} \times |\hat{\mathcal{V}}'|}],
    \begin{aligned}
    ~~~&\text{if}~z=\triangle_{\hat{\mathbf{A}}'},\vartheta>0
    \end{aligned}\\
    &P_{[0,1]}[\mathbf{Y}],~~~~~~~~~~~~~~~~~~~~~~~~z \leq \triangle_{\hat{\mathbf{A}}'}\nonumber
    \end{aligned}
    \right.
\end{align}
where $P_{[0,1]}[\mathbf{Y}]$ maps $\mathbf{Y}$ into $[0,1]$. Following~\cite{feng2022adversarial}, we adopt bisection~\cite{boyd2004convex} to address the equation $z= \triangle_{\hat{\mathbf{A}}'}$ with the dual variable $\vartheta$. Each element is sampled from the Bernoulli distribution as $\mathbf{L}_{\hat{\mathbf{A}}'}[i:j] \sim \text{Bernoulli}(\ddot{\mathbf{L}}_{\hat{\mathbf{A}}'}[i:j])$ to obtain $\mathbf{L}_{\hat{\mathbf{A}}'}$ from $\ddot{\mathbf{L}}_{\hat{\mathbf{A}}'}$.

\subsection{Cross-View Graph Contrastive Learning}
To enhance the ability of \model\ to capture inter-dependencies among different data views and preserve heterogeneous region relations, we introduce a contrastive learning component in addition to our structure-level augmentation over the region graph. We accomplish this by splitting the embedding matrix $\tilde{\textbf{H}}$ into three view-specific matrices $\tilde{\textbf{H}}_p$, $\tilde{\textbf{H}}_m$, and $\tilde{\textbf{H}}_s$, which correspond to three different data views: POI semantic relatedness, urban flow transitions, and geographical locations. To illustrate, let us consider the contrastive loss $\mathcal{L}_{p,m}$ between the POI graph view ($\mathcal{G}_p$, $\tilde{\textbf{H}}_p$) and the mobility graph view ($\mathcal{G}_m$, $\tilde{\textbf{H}}_m$).
\begin{align}
    \label{eq:clLoss}
    \mathcal{L}_{p,m} = \sum_{j\in\mathcal{V}_p\cap\mathcal{V}_m} -\log \frac{\exp({\cos(\tilde{\textbf{h}}_j^p, \tilde{\textbf{h}}_j^m)/\tau})}{ \sum_{j'} \exp({\cos(\tilde{\textbf{h}}_j^p, \tilde{\textbf{h}}_{j'}^m)/\tau})} 
\end{align}
\noindent The remaining two contrastive loss terms, $\mathcal{L}_{m,s}$ ($\mathcal{G}_m$-$\mathcal{G}_s$) and $\mathcal{L}_{p,s}$ ($\mathcal{G}_p$-$\mathcal{G}_s$), can be obtained similarly. To weight these terms, we use $\gamma_{p,m} = \sigma(\text{MLP}(\tilde{\textbf{H}}_p\odot \tilde{\textbf{H}}_m))$, where $\odot$ denotes the element-wise product and $\sigma$ is the ReLU function. The joint cross-view contrastive learning loss is:

\begin{align}
    \label{eq:crossloss}
    \mathcal{L}_\text{cross} = \gamma_{p,m}\cdot \mathcal{L}_{p,m} + \gamma_{m,s} \cdot \mathcal{L}_{m,s} +\gamma_{p,s} \cdot\mathcal{L}_{p,s}
\end{align}

\subsection{Contrastive Model Optimization with InfoMin}
To make our data augmentation adaptive to the contrastive learning tasks, we enhance the reconstruction loss of generative variational autoencoder using a contrast-aware reward function with mutual information minimization~\cite{tian2020makes,wang2019towards}. This function is formally defined:
\begin{align}
\label{eq:infomin}
\text{R}(\tilde{\mathcal{G}}) = \left\{
    \begin{aligned} 
    &1,~~~~~~~\text{if}~ \mathcal{L}_\text{VGAE}>\epsilon\\
    &\xi \ll 1~~\text{otherwise}
    \end{aligned}
    \right.
\end{align}
\noindent where $\epsilon$ is a threshold parameter. To enable robust model training, we incorporate information regularization (IR) into the model as follows:
\begin{align}
    \label{eq:ir}
    &V_\text{IR} = \max(2 \text{s}_1-\text{s}_2 - \text{s}_3, 0);~
    \text{s}_1 = \text{exp}(\frac{\text{cos}(\tilde{\textbf{h}}_{ij}, \tilde{\textbf{h}}'_{ij})}{\tau})\nonumber\\
    &\text{s}_2 = \text{exp}(\frac{\text{cos}(\tilde{\textbf{h}}_{ij}, \textbf{h}_{ij})}{\tau});~
    \text{s}_3 = \text{exp}(\frac{\text{cos}(\tilde{\textbf{h}}'_{ij}, \tilde{\textbf{h}}_{ij})}{\tau})
\end{align}
\noindent where $\tilde{\textbf{H}}'$ denotes the node embedding matrix of $\tilde{\mathcal{G}}'$, which is obtained via Eq~\ref{eq:lowdimGen}. With the reward, the overall optimization objective is to minimize the following loss:
\begin{align}
\label{eq:final_loss}
    \mathcal{L}=\mathcal{L}_\text{VGAE} + \mathcal{L}_\text{cross} + \mathcal{L}_\text{adv} + \mathcal{L}_\text{recon} \cdot \text{R}(\tilde{\mathcal{G}}) + \lambda V_\text{IR}
\end{align}
where $\mathcal{L}_\text{recon}$ denotes the reconstruction loss between the VGAE-derived graph structures $\tilde{G}$ and the original graph $\mathcal{G}$.

\subsection{Model Complexity Analysis}
The \model\ framework incurs computational costs mainly in three aspects: Firstly, the multi-view space-time message passing paradigm has a complexity of $\mathcal{O}(N \times L\times d)$, where $N$ and $L$ denote the number of edges and graph propagation layers, respectively. Secondly, the complexity of our variational adaptive graph augmentation is $\mathcal{O}(|\mathcal{V}|^2\times d)$. Thirdly, for self-supervised loss calculation, \model\ takes $\mathcal{O}(S\times |\tilde{\mathcal{V}}| \times d)$, where $S$ denotes the number of samples included in each batch, and $|\tilde{\mathcal{V}}|$ denotes the average number of nodes in each sampled graph. Despite these costs, our \model\ achieves comparable model efficiency to existing spatial-temporal representation methods. This has been validated in our computational cost evaluation.

\section{Evaluation}
\label{sec:eval}

\begin{table}[t]
\centering
\scriptsize
\setlength{\tabcolsep}{0.1pt}
\caption{Statistics of Experimented Datasets}
\label{fig:data_sta}
\begin{tabular}{c|c|c}
\hline
Dataset        & \textbf{Description of Chicago Data}                                                                                        & \textbf{Description of NYC data} \\ 
\hline 
Blocks  &\begin{tabular}[c]{@{}c@{}}Boundaries of 234 regions split by\\ streets in a certain district, Chicago\end{tabular}         &\begin{tabular}[c]{@{}c@{}}Boundaries of 180 regions split\\ by streets in Manhattan, New York\end{tabular}                             \\ \hline 
Taxi trips & \begin{tabular}[c]{@{}c@{}}Total 386,272 taxi \\trips during a month\end{tabular}                  &\begin{tabular}[c]{@{}c@{}} Total 1,445,285 taxi \\trips during a month\end{tabular}                             \\ \hline
Crimes     &\begin{tabular}[c]{@{}c@{}}Total 321,876 \\2016/01/1-2017/12/30\end{tabular}   &\begin{tabular}[c]{@{}c@{}}Total 108,575 crime\\ 2021/01/1-2022/02/24\end{tabular}                             \\ \hline
POI       &\begin{tabular}[c]{@{}c@{}} Total 3,680,125 POI locations\\ of 130 categories \end{tabular}  &\begin{tabular}[c]{@{}c@{}}Total 20,569 POI locations\\ of 50 categories\end{tabular} 
\\ \hline 
House price       &\begin{tabular}[c]{@{}c@{}}Total 44,447 house price data\\
in a certain district, Chicago\end{tabular}  &\begin{tabular}[c]{@{}c@{}}Total 22,540 house price data\\
in Manhattan, New York\end{tabular} 
\\  \hline
\end{tabular}
\label{tab:detailed_data}
\end{table}

In this section, we evaluate the performance of our \model\ on various spatial-temporal mining tasks, including urban crime forecasting, traffic prediction, and house price prediction. We perform experiments on various datasets, which are described in detail in Table~\ref{tab:detailed_data}. Our goal is to validate the performance of \model\ against state-of-the-art (SOTA) methods on these three tasks.

\subsection{Baselines and Parameter Settings} \vspace{-0.05in}
To evaluate our \model\ method, we consider baselines from three research lines for comparison. The descriptions of the baselines are elaborated in Appendix~\ref{sec:baseline}.

\noindent \textbf{(i) Network Embedding/GNN Models}. We compare \model\ with several representative network embedding and graph neural models, including Node2vec~\cite{grover2016node2vec}, GCN~\cite{kipf2016semi}, GraphSage~\cite{hamilton2017inductive}, GAE~\cite{kipf2016variational}, GAT~\cite{velivckovic2017graph}. We apply these methods to our constructed region graph to generate region embeddings. 


\noindent \textbf{(ii) Graph Contrastive Learning Methods}. In our evaluation, we conduct experimental comparisons between our proposed \model\ method and two recently proposed graph contrastive learning frameworks, GraphCL~\cite{you2020graph} and RGCL~\cite{li2022let}.

\noindent \textbf{(iii) Spatial-Temporal Region Representation}. We further evaluate the performance of our proposed \model\ approach by comparing it with state-of-the-art methods for learning spatial-temporal embeddings of regions, such as POI~\cite{rahmani2019category}, HDGE~\cite{wang2017region}, ZE-Mob~\cite{yao2018representing}, MV-PN~\cite{fu2019efficient}, CGAL~\cite{zhang2019unifying}, MVURE~\cite{zhang2021multi}, and MGFN~\cite{wu2022multi_graph}.

\noindent \textbf{Hyperparameter Settings}. To ensure a fair comparison, we set the dimensionality of the region representation $d$ to 96, which is consistent with the settings used in previous works such as~\cite{zhang2021multi,wu2022multi_graph}. We explore the number of graph propagation layers in the range of \{1,2,3,4,5\} and tune the learning rate to 0.0005 with weight decay of 0.01. We also tune the temperature parameter $\tau$ in the range of \{0.2,0.4,0.6,0.8\}. The baselines are implemented using their original source code. Finally, we tune the weights of the augmented SSL loss in the range of (0,1).


\begin{table*}
\center
\setlength{\abovecaptionskip}{0cm}
\setlength{\belowcaptionskip}{0cm}
\setlength{\tabcolsep}{5pt}
\scriptsize
\caption{Performance comparison in spatial-temporal learning tasks of crime forecasting, traffic prediction, and house price prediction.}
\label{tab:overall}
\begin{tabular}{|c|c|c|c|c|c|c|c|c|c|c|c|c|c|c|}
    \hline
    \multirow{3}{*}{Model} & \multicolumn{4}{c|}{Crime Prediction} & \multicolumn{6}{c|}{Traffic Prediction} & \multicolumn{4}{c|}{House Price Prediction} \\
    \cline{2-15}
    & \multicolumn{2}{c|}{CHI-Crime} & \multicolumn{2}{c|}{NYC-Crime} & \multicolumn{2}{c|}{CHI-Taxi} & \multicolumn{2}{c|}{NYC-Bike} & \multicolumn{2}{c|}{NYC-Taxi} & \multicolumn{2}{c|}{CHI-House} & \multicolumn{2}{c|}{NYC-House}\\ 
    \cline{2-15}
    & MAE & MAPE & MAE & MAPE & MAE & RMSE & MAE & RMSE & MAE & RMSE & MAE & MAPE & MAE & MAPE\\ \cline{2-15} 
    \hline \hline
    Node2vec &1.6334  &0.8605  &4.3646  &0.9454  & \multicolumn{1}{c|}{0.1206} & 0.5803 & \multicolumn{1}{c|}{0.9093} & 1.8513 & \multicolumn{1}{c|}{1.3508} & 4.0105 & \multicolumn{1}{c|}{13137.2178}    &\multicolumn{1}{c|}{44.4278}  & \multicolumn{1}{c|}{4832.6905}    &\multicolumn{1}{c|}{19.8942} \\ 
    \hline
    GCN &1.6061  &0.8546  &4.3257  &0.9234  & \multicolumn{1}{c|}{0.1174} & 0.5707 & \multicolumn{1}{c|}{0.9144} & 1.8321 & \multicolumn{1}{c|}{1.3819} & 4.0200 & \multicolumn{1}{c|}{13074.2121} & \multicolumn{1}{c|}{42.6572} & \multicolumn{1}{c|}{4840.7394} & \multicolumn{1}{c|}{18.3315}\\
    \hline
    GAT &1.5742 &0.8830 &4.3455 &0.9267 & \multicolumn{1}{c|}{0.1105} & 0.5712 & \multicolumn{1}{c|}{0.9110} & 1.8466 & \multicolumn{1}{c|}{1.3746} & 4.0153 & \multicolumn{1}{c|}{13024.7843} &\multicolumn{1}{c|}{43.3221} & \multicolumn{1}{c|}{4799.8482} & 18.3433 \\ 
    \hline
    GraphSage &1.5960 &0.8713 &4.3080 &0.9255 & \multicolumn{1}{c|}{0.1196} & 0.5796 & \multicolumn{1}{c|}{0.9102} & 1.8473 & \multicolumn{1}{c|}{1.3966} & 4.0801 & \multicolumn{1}{c|}{13145.5623} & \multicolumn{1}{c|}{44.3167} & \multicolumn{1}{c|}{4875.6026} & \multicolumn{1}{c|}{18.4570} \\ 
    \hline
    GAE &1.5711 &0.8801 &4.3749 &0.9343 & \multicolumn{1}{c|}{0.1103} & 0.5701 & \multicolumn{1}{c|}{0.9132} & 1.8412 & \multicolumn{1}{c|}{1.3719} & 4.0337 & \multicolumn{1}{c|}{13278.3256} & \multicolumn{1}{c|}{42.3221} & \multicolumn{1}{c|}{4896.9564} & \multicolumn{1}{c|}{18.3114}\\
    \hline
    GraphCL &1.2332 &0.6293 &3.3075 &0.6771 & \multicolumn{1}{c|}{0.0812} &0.5364  & \multicolumn{1}{c|}{0.8582} &1.8180  & \multicolumn{1}{c|}{1.3022} &3.7029  & \multicolumn{1}{c|}{10752.5693} & \multicolumn{1}{c|}{28.8374} & \multicolumn{1}{c|}{4562.7279} & \multicolumn{1}{c|}{11.3055}\\
    \hline
    RGCL &1.1946 &0.6081 &3.1026 &0.6323 & \multicolumn{1}{c|}{0.0797} &0.5052  & \multicolumn{1}{c|}{0.8319} &1.8107  & \multicolumn{1}{c|}{1.2973} &3.6865  & \multicolumn{1}{c|}{10673.3289} & \multicolumn{1}{c|}{27.5279} & \multicolumn{1}{c|}{4439.0733} & \multicolumn{1}{c|}{10.2375}\\
    \hline
    POI &1.3047 &0.8142 &4.0069 &0.8658 & \multicolumn{1}{c|}{0.0933} & 0.5578 & \multicolumn{1}{c|}{0.8892} & 1.8277 & \multicolumn{1}{c|}{1.3316} & 3.9872 & \multicolumn{1}{c|}{12045.3212} & \multicolumn{1}{c|}{33.5049} & \multicolumn{1}{c|}{4703.3755} & \multicolumn{1}{c|}{16.7920}\\ 
    \hline
    HDGE &1.3586 &0.8273 &4.2021 &0.7821 & \multicolumn{1}{c|}{0.0865} & 0.5502 & \multicolumn{1}{c|}{0.8667} & 1.8251 & \multicolumn{1}{c|}{1.2997} & 3.9846 & \multicolumn{1}{c|}{11976.3215} & \multicolumn{1}{c|}{30.8451} & \multicolumn{1}{c|}{4677.6905} & \multicolumn{1}{c|}{12.5192}\\ 
    \hline
    ZE-Mob &1.3954 &0.8249 &4.3560 &0.8012 & \multicolumn{1}{c|}{0.1002} & 0.5668 & \multicolumn{1}{c|}{0.8900} & 1.8359 & \multicolumn{1}{c|}{1.3314} & 4.0366 & \multicolumn{1}{c|}{12351.1321} & \multicolumn{1}{c|}{38.6171} & \multicolumn{1}{c|}{4730.6927} & \multicolumn{1}{c|}{16.2586}\\ 
    \hline
    MV-PN &1.3370 &0.8132 &4.2342 &0.7791 & \multicolumn{1}{c|}{0.0903} & 0.5502 & \multicolumn{1}{c|}{0.8886} & 1.8313 & \multicolumn{1}{c|}{1.3306} & 3.9530 & \multicolumn{1}{c|}{12565.0607} & \multicolumn{1}{c|}{39.7812} & \multicolumn{1}{c|}{4798.2951} & \multicolumn{1}{c|}{17.0418}\\ 
    \hline
    CGAL &1.3386 &0.7950 &4.1782 &0.7506 & \multicolumn{1}{c|}{0.1013} & 0.5682 & \multicolumn{1}{c|}{0.9097} & 1.8557 & \multicolumn{1}{c|}{1.3353} & 4.0671 & \multicolumn{1}{c|}{12094.5869}    &\multicolumn{1}{c|}{36.9078}     & \multicolumn{1}{c|}{4731.8159}    &\multicolumn{1}{c|}{16.5454}\\
    \hline
    MVURE &1.2586 &0.7087 &3.7683 &0.7318 & \multicolumn{1}{c|}{0.0874} & 0.5405 & \multicolumn{1}{c|}{0.8699} & 1.8157 & \multicolumn{1}{c|}{1.3007} & 3.6715 & \multicolumn{1}{c|}{11095.5323}    &\multicolumn{1}{c|}{34.8954}     & \multicolumn{1}{c|}{4675.1626}    &\multicolumn{1}{c|}{15.9860}\\ 
    \hline
    MGFN &1.2538 &0.6937 &3.5971 &0.7065 & \multicolumn{1}{c|}{0.0831} & 0.5385 & \multicolumn{1}{c|}{0.8783} & 1.8163 & \multicolumn{1}{c|}{1.3266} & 3.7514 & \multicolumn{1}{c|}{10792.7834}    &\multicolumn{1}{c|}{29.9832}    & \multicolumn{1}{c|}{4651.3451}    &\multicolumn{1}{c|}{12.9752}\\ 
    \hline \hline 
    \emph{Ours} & \textbf{1.1285} & \textbf{0.5740} & \textbf{2.1060} & \textbf{0.5203} & \textbf{0.0747} & \textbf{0.4938} & \textbf{0.7787} & \textbf{1.8079} & \textbf{1.2892} & \textbf{3.6616} & \textbf{10461.5321} & \textbf{26.5092} & \textbf{4347.9815} & \textbf{7.0510}\\
    \hline
\end{tabular}
\vspace{-1.0em}
\end{table*}

\subsection{Urban Crime Forecasting}
\noindent \textbf{Setup}. The urban crime forecasting task aims to predict the number of crime events that will occur in future time slots. We conduct experiments on two real-world crime datasets collected from Chicago (2018) and New York City (2019). Since our \model\ is model-agnostic for downstream tasks, we utilize it as a pre-trained module to integrate with the state-of-the-art crime prediction model ST-SHN~\cite{xia21spatial}. The encoded region embeddings from \model\ are then fine-tuned using ST-SHN. We follow the same settings as ST-SHN, including the region partition strategy (Chicago: 234, NYC: 180 regions), training/test data split, and evaluation metrics (MAE and MAPE). To conserve space, we reported the overall prediction accuracy across different crime categories, such as Theft, Assault, and Robbery. Category-specific prediction results and detailed settings are included in the supplementary material.\\\vspace{-0.12in}

\noindent \textbf{Results}. Based on the reported results in Table~\ref{tab:overall}, it is evident that our proposed \model\ outperforms state-of-the-art spatial-temporal region representation methods in all cases. This significant improvement in performance validates the effectiveness of our approach in addressing the challenges posed by sparse and skewed crime data. Our model achieves this by providing effective self-augmented signals for the region representation paradigm. Furthermore, applying standard graph neural networks directly over the region graph can be susceptible to noisy spatial-temporal region relations generated from heterogeneous data sources. This vulnerability can result in suboptimal representations.

\subsection{Traffic Flow Prediction}
\noindent \textbf{Setup}. To assess the effectiveness of our proposed \model\ in predicting citywide traffic volume, we evaluated all compared methods on the backbone of ST-GCN~\cite{yu2017spatio}. The results, evaluated in terms of MAE and MAPE on three traffic benchmark datasets, are reported in Table~\ref{tab:overall}. Following the settings in ST-GCN, we set the historical time periods to 60 minutes with 12 traffic volume records to make predictions for the next 15, 30, and 45 minutes.\\\vspace{-0.12in}

\noindent \textbf{Results}. Our experimental results demonstrate that our proposed \model\ consistently outperforms all baselines on different datasets, indicating its effectiveness in capturing region-wise traffic dependencies with noisy region graph connections. This is in contrast to current region embedding models, which are limited in accurately capturing such dependencies, particularly in spatially adjacent areas with different urban functions (POI semantic relatedness). Furthermore, while some methods, such as MV-PN, CGAL, and MGFN, attempt to consider mobility data for region correlation modeling, the skewed mobility and traffic distribution make them easily biased towards regions with dense data, sacrificing the representation performance of unpopular spatial areas. This highlights the advantages of our proposed \model\ in addressing these limitations and improve the quality of spatial-temporal graph representation learning.

\subsection{House Price Prediction}

\noindent \textbf{Setup}. To further evaluate the effectiveness of \model, we conduct experiments to predict region-specific house prices using data collected from 22,540 houses in NYC and 44,447 houses in Chicago. We follow the same house price mapping method as in~\cite{wang2017region}. The region embeddings learned by different methods are then fed into Lasso regression to make house price predictions.\\\vspace{-0.12in}

\noindent \textbf{Results}. Our experimental results demonstrate that our \model\ outperforms all baselines in the house price prediction task, confirming its superior region representation capacity. Benefiting from our adversarial adaptive graph augmentation, \model\ can preserve the holistic semantics across time and space in the latent region embedding space, despite noisy and incomplete multi-view spatial-temporal data. This is an advantage over previous approaches, which fail to effectively capture such dependencies

\subsection{Ablation Study and Effectiveness Analyses}
In this section, we investigate the impact of each component of \model\ on model performance by constructing four variants of the model. Our goal is to answer the question: \emph{Is it important to incorporate \model's key components in boosting the model performance?} The five variants include: (1) ``w/o adversarial view (ADS)'' which disables the adversarial learning component against hard samples; (2) ``w/o VGAE'' which replaces VGAE with random edge dropout operators; (3) ``w/o information regularization (IR)'' which disables \model's ability to identify hard negative samples; (4) ``w/o information minimization (IM)'' which removes mutual information minimization for redundant information alleviation. By comparing the performance of these variants with the original \model, we aim to determine the importance of each component in improving performance.

\begin{figure}
\centering
\begin{tabular}{c c c c }
\hspace{25.5mm}
\begin{minipage}{0.5cm}
\includegraphics[width=3.5cm]{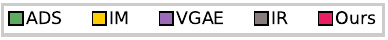}
\end{minipage}\vspace{0.5mm}\hspace{-18.5mm}
&
\\
\hspace{-5.5mm}
  \begin{minipage}{0.115\textwidth}
	\includegraphics[width=\textwidth]{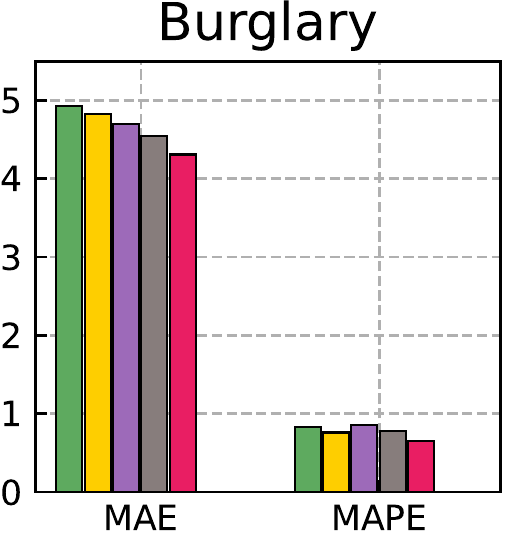}
  \end{minipage}\hspace{-3.5mm}
  &
  \begin{minipage}{0.12\textwidth}
    \includegraphics[width=\textwidth]{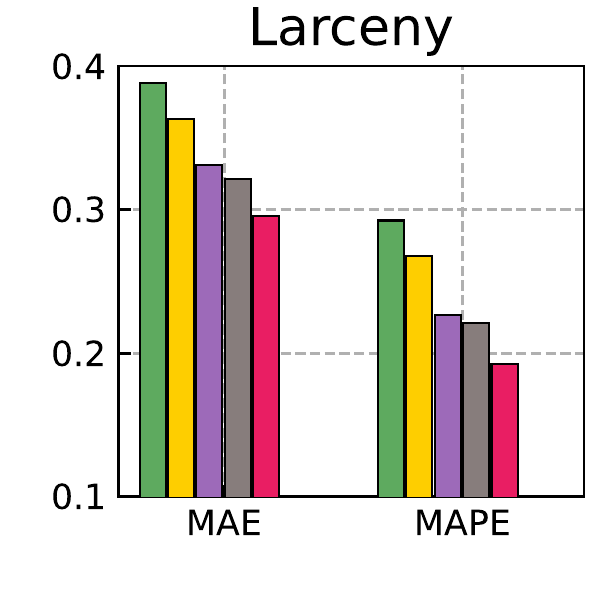}
  \end{minipage}\hspace{-3.5mm}
  &
  \begin{minipage}{0.12\textwidth}
	\includegraphics[width=\textwidth]{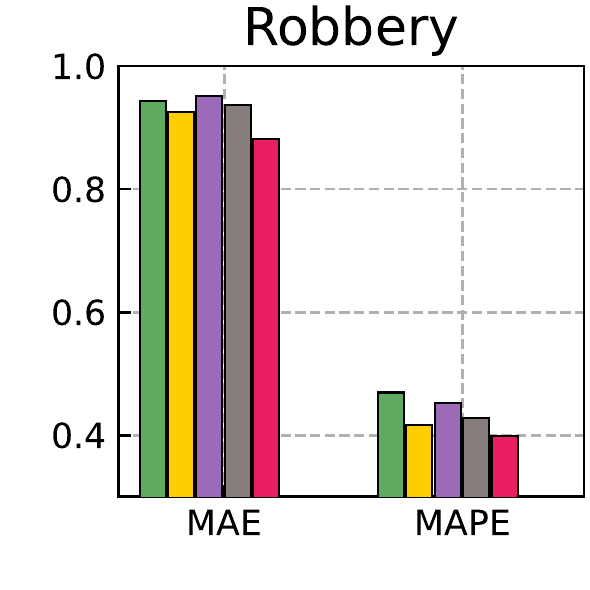}
  \end{minipage}\hspace{-3.5mm}
  &
  \begin{minipage}{0.12\textwidth}
    \includegraphics[width=\textwidth]{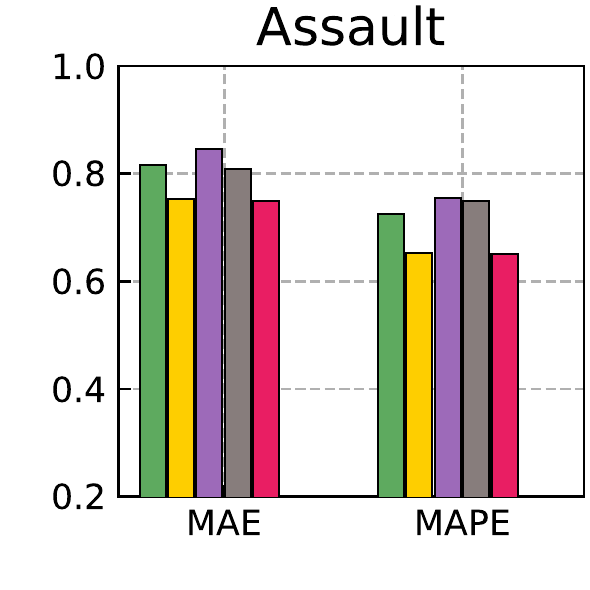}
  \end{minipage}\hspace{-3.5mm}
\end{tabular}
\vspace{-0.12in}
\caption{Ablation study of \model\ for crime prediction.}
\vspace{-0.15in}
\label{fig:ablation}
\end{figure}

We report the category-specific crime forecasting results in Figure~\ref{fig:ablation} and traffic prediction results on NYC-Taxi and NYC-bike in Table~\ref{fig:ablation_traffic}. Our results demonstrate that all key components of \model\ contribute significantly to learning expressive and robust region representations, as evidenced by the improved performance. Furthermore, some simplified \model\ variants that include only the adversarial adaptive augmentation achieve promising performance, indicating the effectiveness of the region representation with multi-view graph self-supervised augmentation underlying spatial-temporal dynamics for urban computing applications. 



\begin{table}[t]
\center
\setlength{\abovecaptionskip}{-0mm}
\setlength{\belowcaptionskip}{0cm}
\setlength{\tabcolsep}{8.5pt}
\footnotesize
\caption{Ablation study of \model\ for traffic prediction}
\label{fig:ablation_traffic}
\begin{tabular}{|c|cc|cc|}
\hline
\multirow{2}{*}{Model} & \multicolumn{2}{c|}{NYC-Taxi}   & \multicolumn{2}{c|}{NYC-Bike}   \\ \cline{2-5} 
                       & \multicolumn{1}{c|}{MAE} & RMSE & \multicolumn{1}{c|}{MAE} & RMSE \\ \hline
w/o ADS                     & \multicolumn{1}{c|}{1.3731}    &3.8764      & \multicolumn{1}{c|}{0.9067}    &1.9516      \\ \hline
w/o IM                   & \multicolumn{1}{c|}{1.3519}    &3.8533      & \multicolumn{1}{c|}{0.9132}    &1.9304      \\ \hline
w/o VGAE                   & \multicolumn{1}{c|}{1.3486}    &3.8457      & \multicolumn{1}{c|}{0.8941}    &1.9072      \\ \hline
w/o IR                   & \multicolumn{1}{c|}{1.3418}    &3.7904      & \multicolumn{1}{c|}{0.8837}    &1.8705      \\ \hline
Ours                   & \multicolumn{1}{c|}{\textbf{1.2892}}    &\textbf{3.6616}      & \multicolumn{1}{c|}{\textbf{0.7787}}    &\textbf{1.8079}      \\ \hline
\end{tabular}
\vspace{-0.1in}
\end{table}

\subsection{Performance over Sparse Data}
\begin{figure}[h]
\vspace*{-1mm}
\centering
\begin{tabular}{c c c c}
\hspace{-50.5mm}
\begin{minipage}{0.5cm}
\includegraphics[width=7.5cm]{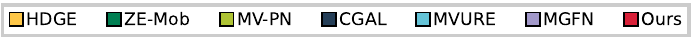}
\end{minipage}\vspace{0.5mm}\hspace{-50.5mm}
&
\\
\hspace{-5.5mm}
  \begin{minipage}{0.12\textwidth}
    \includegraphics[width=\textwidth]{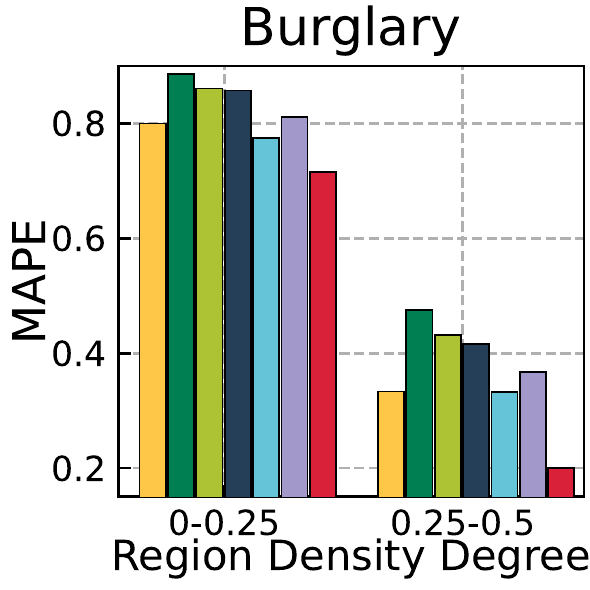}
  \end{minipage}\hspace{-3.5mm}
  &
  \begin{minipage}{0.12\textwidth}
    \includegraphics[width=\textwidth]{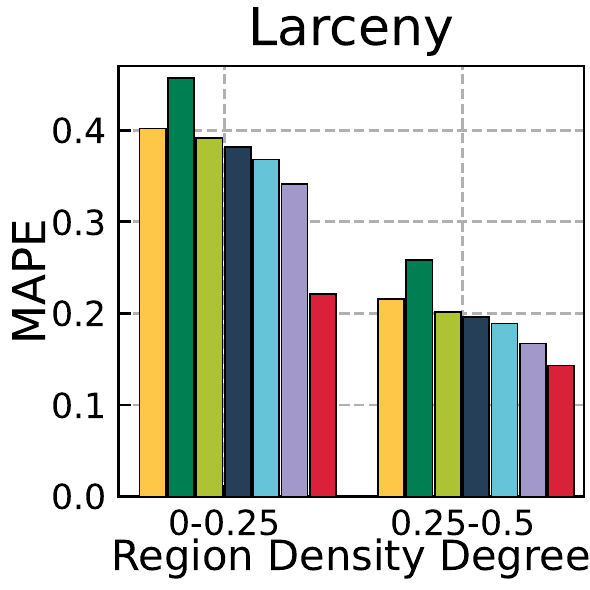}
  \end{minipage}\hspace{-3.5mm}
  &
  \begin{minipage}{0.12\textwidth}
    \includegraphics[width=\textwidth]{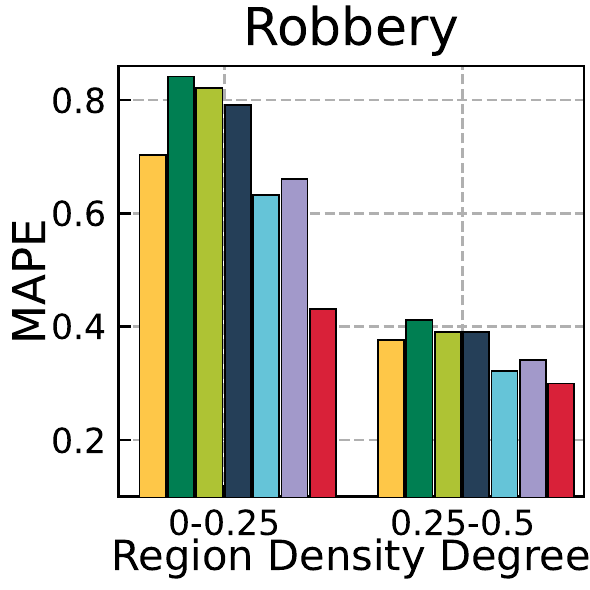}
  \end{minipage}\hspace{-3.5mm}
  &
  \begin{minipage}{0.12\textwidth}
    \includegraphics[width=\textwidth]{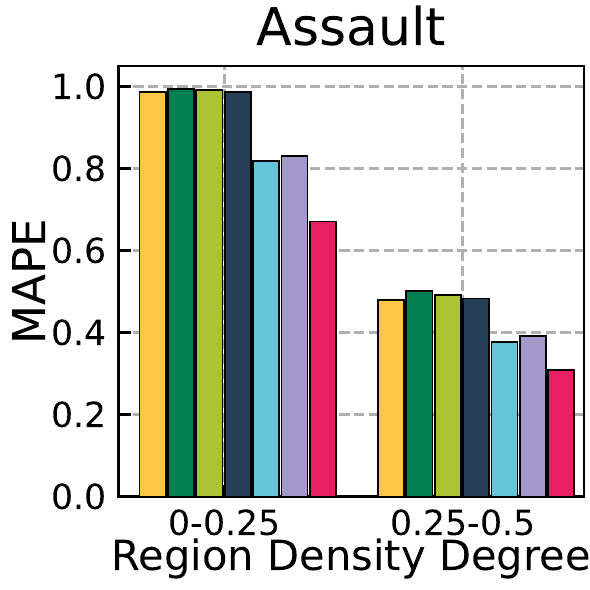}
  \end{minipage}\hspace{-3.5mm}
\end{tabular}
\vspace{-0.1in}
\caption{Evaluation Results on New York Data for different criminal offense types \wrt\ different data density degrees.}
\vspace*{-2mm}
\label{fig:robustness}
\end{figure}

In this section, we investigate the robustness of \model\ in alleviating the effects of data sparsity. Specifically, we evaluate the representation performance on regions with different crime data density degrees, ranging from $(0.0, 2.5]$ to $(2.5, 5.0]$, based on the non-zero value ratio of region-specific crime sequences. The results, presented in Figure~\ref{fig:robustness}, demonstrate that our \model\ consistently improves performance under different degrees of data sparsity. This can be attributed to the self-supervision signals that are derived from the intrinsic structure of multi-view spatial-temporal data, which allows \model\ to learn effective representations even in regions with sparse crime data.

\subsection{Effects of Hyperparameters}
\begin{figure}
\centering
\begin{tabular}{c c c }
\hspace{-5.5mm}
  \begin{minipage}{0.15\textwidth}
	\includegraphics[width=\textwidth]{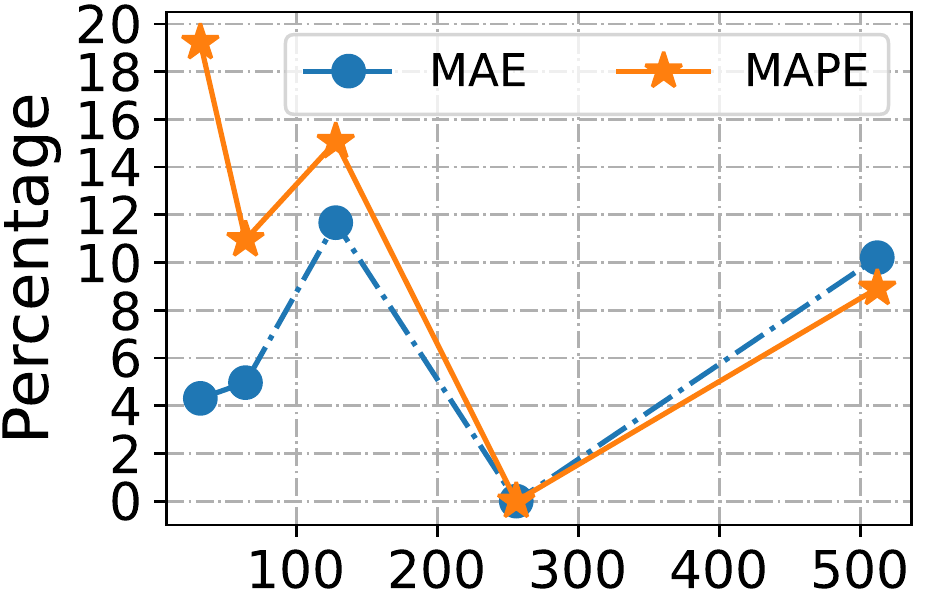}
  \end{minipage}\hspace{-3.5mm}
  &
  \begin{minipage}{0.15\textwidth}
    \includegraphics[width=\textwidth]{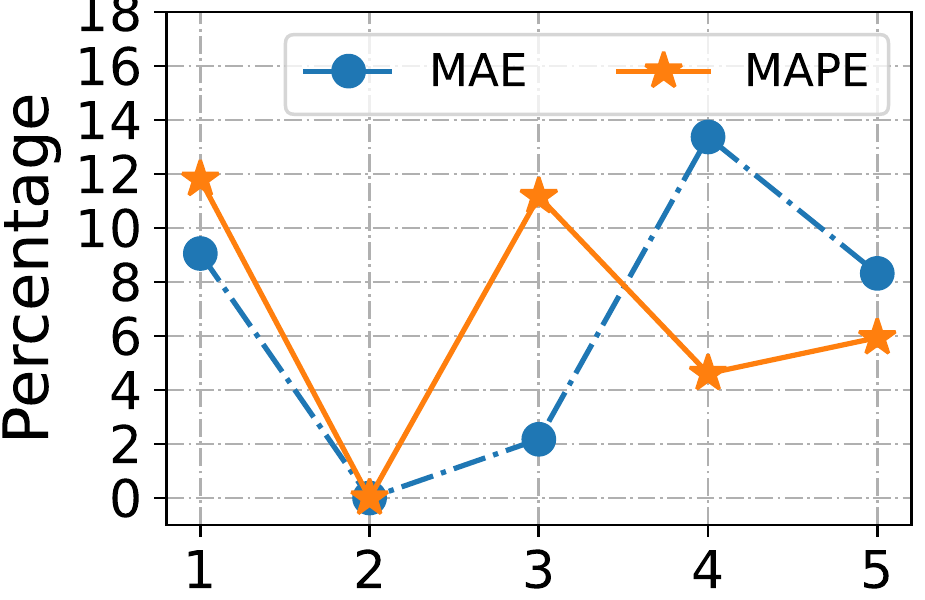}
  \end{minipage}\hspace{-3.5mm}
  &
  \begin{minipage}{0.15\textwidth}
	\includegraphics[width=\textwidth]{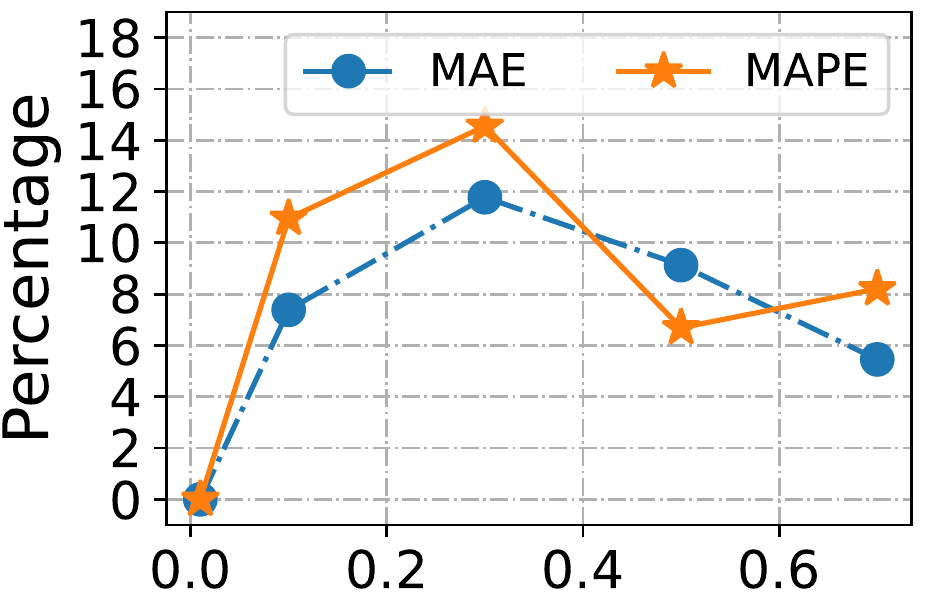}
  \end{minipage}\hspace{-3.5mm}
  \\
 (a) \# Hidden
 &
 \hspace{6.5mm}
 (b) \# Layer 
 &
 \hspace{4.5mm}
 (c) $\xi$
\end{tabular}
\vspace{-0.12in}
\caption{Hyperparameter effects of \model\ on NYC crime data.}
\vspace{-0.3in}
\label{fig:hyper}
\end{figure}

We investigate the effects and sensitivity of our proposed \model\ to different hyperparameters. Figure~\ref{fig:hyper} shows the percentage of performance degradation compared to the best performance when varying each hyperparameter, while keeping the other parameters at their default values. Our results show that increasing the embedding dimensionality beyond a certain point (tuning range \{$2^5$, $2^6$, $2^7$, $2^8$, $2^9$\}) leads to more noise and overfitting, resulting in degraded performance. Similarly, increasing the number of GNN layers beyond a certain point (tuned from \{1,2,3,4,5\}) can lead to over-smoothing and indistinguishable representations, which also results in degraded performance. Finally, we find that the best performance is achieved with a value of $\xi=0.01$ from \{0.01,0.1,0.3,0.5,0.7\}.


\subsection{Case Study}
\begin{figure}
    \centering
    \includegraphics[width=0.94\columnwidth]{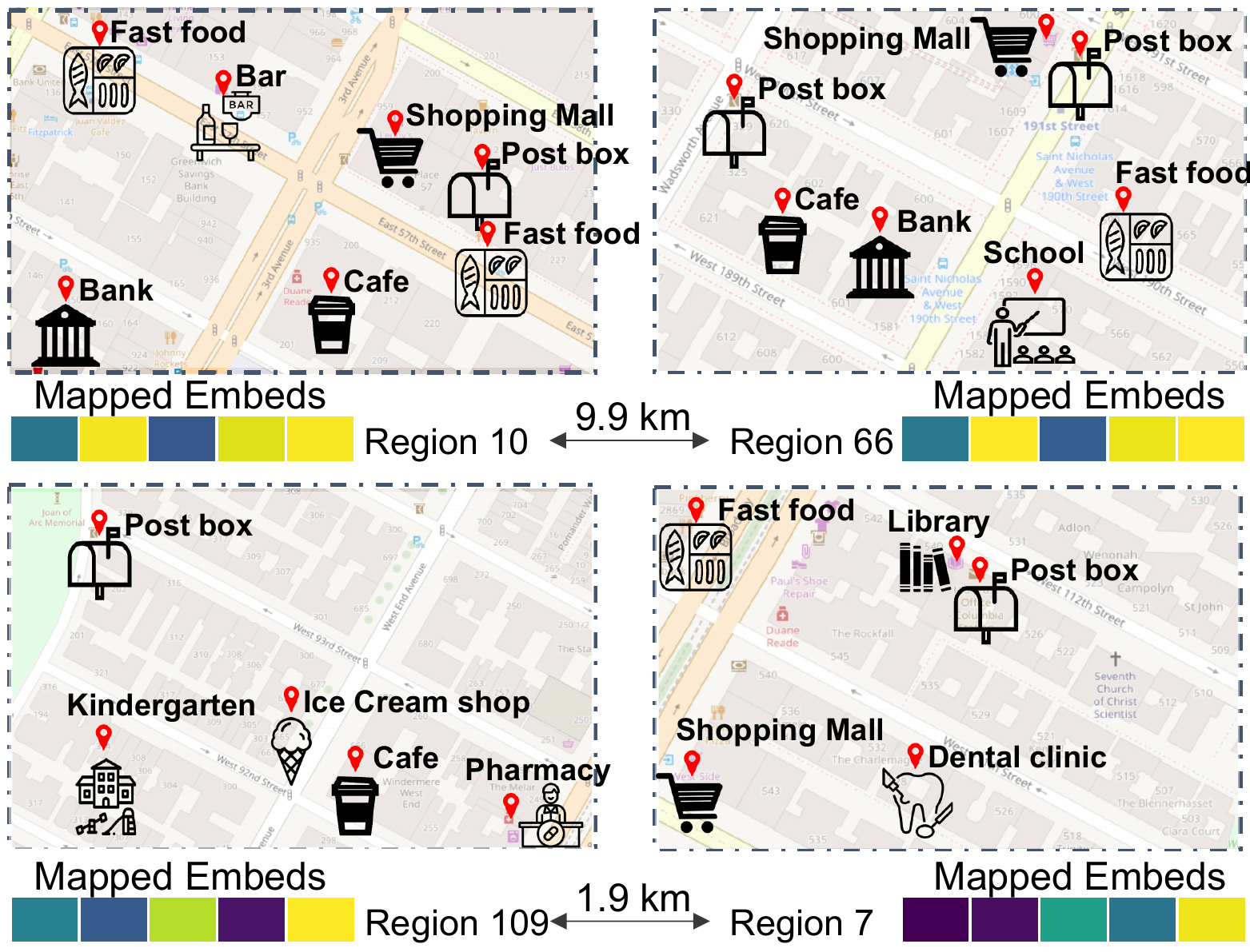}
    \vspace{-0.05in}
    \caption{Case study of \model\ on New York City. Our \model\ can alleviate the connection bias in spatial-temporal graphs by i) refining the salient relational signals between far away regions (region 10 \& region 66) ii) denoising connections between adjacent regions with weak dependence (region 7 \& region 109).}
    \label{fig:my_label}
    \vspace{-0.15in}
\end{figure}

This section presents case studies to demonstrate the effectiveness of our proposed \model\ in modeling cross-region dependencies and distilling global spatial-temporal knowledge while alleviating view-specific noise perturbation. Figure~\ref{fig:my_label} visualizes the region embeddings mapped from the original representations encoded by \model. We observe that the long distance between two functionally similar regions, $r_{10}$ and $r_{66}$, may lead to weak connections in the multi-view region graph. However, our adversarial self-supervised augmentation paradigm enables \model\ to distill the implicit region dependencies from the noisy information. Furthermore, while regions $r_7$ and $r_{109}$ exhibit different function semantics in the urban space, their spatially adjacent relationship results in direct connections over the graph structure. Our \model\ can overcome this bias by avoiding misleading the graph neural network with heavy propagation between these less relevant regions.


\subsection{Model Efficiency Study}
\label{sec:efficiency}
Finally, we investigate the model efficiency of \model\ by comparing it with state-of-the-art region representation methods in terms of training time, as shown in Table~\ref{fig:efficiency}. All methods are implemented in Python 3.8, PyTorch 1.7.0 (GPU version), and TensorFlow 1.15.3 (GPU version) (ST-SHN). The experiments are conducted on a server with 10 cores of Intel(R) Core(TM) i9-9820X CPU @ 3.30GHz, 64.0GB RAM, and one Nvidia GeForce RTX 3090 GPU. We observe that \model\ achieves competitive model efficiency compared to the baselines, demonstrating its potential in handling large-scale spatial-temporal data. The augmented self-supervised learning paradigms do not involve much additional computational cost but significantly boost the region representation performance. 


\begin{table}[h]
\center
\setlength{\tabcolsep}{1.0pt}
\footnotesize
\caption{Computational cost (seconds) of our \model\ and SOTA spatial-temporal region representation methods. MAE and MAPE are estimated for NYC crime prediction results.}
\label{fig:efficiency}
\begin{tabular}{c|c|c|c|c|c|c|c}
\hline
Models & HDGE & ZE-Mob & MV-PN & CGAL & MVURE & MGFN & Ours \\ \hline
Training        &303.7     &82.7       &33.4      &4144.8       &240.7       &852.3     &275.1     \\\hline
MAE        &4.2021     &4.3560        &4.2342       &4.1782       &3.7683       &3.5971      &2.1060      \\\hline
MAPE        &0.7821     &0.8012        &0.7791       &0.7506       &0.7318       &0.7065      &0.5203      \\\hline
\end{tabular}
\end{table}

\section{Related Work}
\label{sec:relate}

\subsection{Spatial-Temporal Region Representation Learning}
Many efforts have been devoted to proposing various region representation solutions based on spatial-temporal data. For example, an early study~\cite{wang2017region} utilized network embedding techniques to map region networks into latent representations. Motivated by the effectiveness of graph neural network for relational learning, many GNN-based models are introduced to perform the message passing over the generated region graph structures for region embedding refinement by considering the correlations between connected regions~\cite{zhang2021multi,zhang2019unifying,fu2019efficient,wu2022multi_graph}. However, most of the existing models are vulnerable to the quality of the constructed region graphs. In response to this limitation, this work proposes a self-supervised learning paradigm that explores data augmentation adaptive to spatial-temporal region dependencies. \\\vspace{-0.12in}

\subsection{Graph Neural Networks for Spatial-Temporal Data}
In recent years, Graph Neural Networks (GNNs) have been successfully applied to representation learning on various spatial-temporal data. One notable example is traffic modeling, where GNNs are used to propagate information among neighboring geographical regions for traffic pattern representations~\cite{li2018diffusion,zhang2021traffic,li2021spatial}. While some attempts propose to discriminate the region-wise message passing with graph attention networks~\cite{zheng2020gman,fang2021ftpg}, the noisy graph structures can still mislead the derivation of attentive weights. To encode high-order region correlations, GNN-enhanced crime predictive models are developed to encode crime dynamics~\cite{xia21spatial,wang2022hagen}. Additionally, the location recommender systems are improved by graph neural models to jointly encode information from both temporal and spatial dimensions~\cite{yang2022getnext,lim2022hierarchical}. The use of graph contrastive learning in embedding spatial-temporal data has shown promising results in recent attempts~\cite{zhang2023automated}. Our new framework takes a step further by explicitly incorporating adversarial SSL and cross-view CL paradigms to enhance spatial-temporal graph learning. With our framework, we provide a powerful and flexible solution for embedding spatial-temporal data.


\subsection{Graph Contrastive Learning}\vspace{-0.05in}

Recent studies have shown the benefits of contrastive learning in graph representations to address the data sparsity and noise challenges~\cite{zhu2021graph,xia2023graph,lin2022prototypical,yin2022autogcl} by generating contrastive views with various techniques. These studies generate contrastive views using various techniques, such as stochastic node/edge masking strategies~\cite{you2020graph}. Follow-up models have been proposed to improve the representation power of graph contrastive learning, including rationale-aware contrastive augmentation~\cite{li2022let}, exploring information transformation in GCL~\cite{xu2021infogcl}, and latent factor disentanglement~\cite{li2021disentangled}. Motivated by those approaches, the proposed \model\ method has the potential to offer significant benefits for spatial-temporal graph learning tasks by providing effective SSL signal distillation.



\section{Conclusion}
\label{sec:conclusion}

In this paper, we introduce the \model\ model for spatial-temporal graph learning, which enables self-supervised augmentation over spatial-temporal graphs. Our model designs an adversarial contrastive learning paradigm with generative autoencoder. We also propose a cross-view contrastive learning method that takes into account the heterogeneity of region relation to improve inter-dependency modeling. Our experimental results on various spatial-temporal prediction tasks demonstrate the superior performance of our method. Our distillation of essential self-supervised information also reduces the need for manual feature engineering.



\vspace{-0.1cm}

\section*{Acknowledgments}
This project is partially supported by HKU-SCF FinTech Academy and Shenzhen-Hong Kong-Macao Science and Technology Plan Project (Category C Project: SGDX20210823103537030) and Theme-based Research Scheme T35-710/20-R. We also thank the Department of Computer Science and the Musketeers Foundation Institute of Data Science at HKU for their support in this work.

\clearpage

\nocite{langley00}

\balance
\bibliography{refs_full}

\begin{thebibliography}{46}
\providecommand{\natexlab}[1]{#1}
\providecommand{\url}[1]{\texttt{#1}}
\expandafter\ifx\csname urlstyle\endcsname\relax
  \providecommand{\doi}[1]{doi: #1}\else
  \providecommand{\doi}{doi: \begingroup \urlstyle{rm}\Url}\fi

\bibitem[Boyd et~al.(2004)Boyd, Boyd, and Vandenberghe]{boyd2004convex}
Boyd, S., Boyd, S.~P., and Vandenberghe, L.
\newblock \emph{Convex optimization}.
\newblock Cambridge university press, 2004.

\bibitem[Fang et~al.(2021)Fang, Tang, Yang, Chen, Li, and Li]{fang2021ftpg}
Fang, M., Tang, L., Yang, X., Chen, Y., Li, C., and Li, Q.
\newblock Ftpg: A fine-grained traffic prediction method with graph attention
  network using big trace data.
\newblock \emph{Transactions on Intelligent Transportation Systems (TITS)},
  2021.

\bibitem[Feng et~al.(2019)Feng, Zhang, Wang, Yang, Zhang, Li, and
  Jin]{feng2019dplink}
Feng, J., Zhang, M., Wang, H., Yang, Z., Zhang, C., Li, Y., and Jin, D.
\newblock Dplink: User identity linkage via deep neural network from
  heterogeneous mobility data.
\newblock In \emph{The Web Conference (WWW)}, pp.\  459--469, 2019.

\bibitem[Feng et~al.(2022)Feng, Jing, Zhu, and Tong]{feng2022adversarial}
Feng, S., Jing, B., Zhu, Y., and Tong, H.
\newblock Adversarial graph contrastive learning with information
  regularization.
\newblock In \emph{The Web Conference (WWW)}, pp.\  1362--1371, 2022.

\bibitem[Fu et~al.(2019)Fu, Wang, Du, Wu, and Li]{fu2019efficient}
Fu, Y., Wang, P., Du, J., Wu, L., and Li, X.
\newblock Efficient region embedding with multi-view spatial networks: A
  perspective of locality-constrained spatial autocorrelations.
\newblock In \emph{International Conference on Artificial Intelligence (AAAI)},
  volume~33, pp.\  906--913, 2019.

\bibitem[Grover \& Leskovec(2016)Grover and Leskovec]{grover2016node2vec}
Grover, A. and Leskovec, J.
\newblock node2vec: Scalable feature learning for networks.
\newblock In \emph{International Conference on Knowledge Discovery and Data
  Mining (KDD)}, pp.\  855--864, 2016.

\bibitem[Hamilton et~al.(2017)Hamilton, Ying, and
  Leskovec]{hamilton2017inductive}
Hamilton, W., Ying, Z., and Leskovec, J.
\newblock Inductive representation learning on large graphs.
\newblock In \emph{International Conference on Neural Information Processing
  Systems (NeurIPS)}, volume~30, 2017.

\bibitem[Jin et~al.(2020)Jin, Ma, Liu, Tang, Wang, and Tang]{jin2020graph}
Jin, W., Ma, Y., Liu, X., Tang, X., Wang, S., and Tang, J.
\newblock Graph structure learning for robust graph neural networks.
\newblock In \emph{International Conference on Knowledge Discovery and Data
  Mining (KDD)}, pp.\  66--74, 2020.

\bibitem[Kipf \& Welling(2016)Kipf and Welling]{kipf2016variational}
Kipf, T.~N. and Welling, M.
\newblock Variational graph auto-encoders.
\newblock \emph{arXiv preprint arXiv:1611.07308}, 2016.

\bibitem[Kipf \& Welling(2017)Kipf and Welling]{kipf2016semi}
Kipf, T.~N. and Welling, M.
\newblock Semi-supervised classification with graph convolutional networks.
\newblock In \emph{International Conference on Learning Representations
  (ICLR)}, 2017.

\bibitem[Li et~al.(2021)Li, Wang, Zhang, Yuan, Li, and Zhu]{li2021disentangled}
Li, H., Wang, X., Zhang, Z., Yuan, Z., Li, H., and Zhu, W.
\newblock Disentangled contrastive learning on graphs.
\newblock In \emph{International Conference on Neural Information Processing
  Systems (NeurIPS)}, volume~34, pp.\  21872--21884, 2021.

\bibitem[Li \& Zhu(2021)Li and Zhu]{li2021spatial}
Li, M. and Zhu, Z.
\newblock Spatial-temporal fusion graph neural networks for traffic flow
  forecasting.
\newblock In \emph{International Conference on Artificial Intelligence (AAAI)},
  pp.\  4189--4196, 2021.

\bibitem[Li et~al.(2022{\natexlab{a}})Li, Wang, Zhang, Wu, He, and
  Chua]{li2022let}
Li, S., Wang, X., Zhang, A., Wu, Y., He, X., and Chua, T.-S.
\newblock Let invariant rationale discovery inspire graph contrastive learning.
\newblock In \emph{International Conference on Machine Learning (ICML)}, pp.\
  13052--13065. PMLR, 2022{\natexlab{a}}.

\bibitem[Li et~al.(2018)Li, Yu, Shahabi, and Liu]{li2018diffusion}
Li, Y., Yu, R., Shahabi, C., and Liu, Y.
\newblock Diffusion convolutional recurrent neural network: Data-driven traffic
  forecasting.
\newblock In \emph{International Conference on Learning Representations
  (ICLR)}, 2018.

\bibitem[Li et~al.(2022{\natexlab{b}})Li, Huang, Xia, Xu, and
  Pei]{li2022spatial}
Li, Z., Huang, C., Xia, L., Xu, Y., and Pei, J.
\newblock Spatial-temporal hypergraph self-supervised learning for crime
  prediction.
\newblock In \emph{International Conference on Data Engineering (ICDE)}, pp.\
  2984--2996. IEEE, 2022{\natexlab{b}}.

\bibitem[Lim et~al.(2022)Lim, Hooi, Ng, Goh, Weng, and
  Tan]{lim2022hierarchical}
Lim, N., Hooi, B., Ng, S.-K., Goh, Y.~L., Weng, R., and Tan, R.
\newblock Hierarchical multi-task graph recurrent network for next poi
  recommendation.
\newblock In \emph{International Conference on Research and Development in
  Information Retrieval (SIGIR)}, 2022.

\bibitem[Lin et~al.(2022)Lin, Liu, Zhou, Hu, Wang, Zhao, Zheng, Lin, Xing, and
  Liang]{lin2022prototypical}
Lin, S., Liu, C., Zhou, P., Hu, Z.-Y., Wang, S., Zhao, R., Zheng, Y., Lin, L.,
  Xing, E., and Liang, X.
\newblock Prototypical graph contrastive learning.
\newblock \emph{Transactions on Neural Networks and Learning Systems (TNNLS)},
  2022.

\bibitem[Pan et~al.(2019)Pan, Liang, Wang, Yu, Zheng, and Zhang]{pan2019urban}
Pan, Z., Liang, Y., Wang, W., Yu, Y., Zheng, Y., and Zhang, J.
\newblock Urban traffic prediction from spatio-temporal data using deep meta
  learning.
\newblock In \emph{International Conference on Knowledge Discovery and Data
  Mining (KDD)}, pp.\  1720--1730, 2019.

\bibitem[Rahmani et~al.(2019)Rahmani, Aliannejadi, Mirzaei~Zadeh, Baratchi,
  Afsharchi, and Crestani]{rahmani2019category}
Rahmani, H.~A., Aliannejadi, M., Mirzaei~Zadeh, R., Baratchi, M., Afsharchi,
  M., and Crestani, F.
\newblock Category-aware location embedding for point-of-interest
  recommendation.
\newblock In \emph{International Conference on Research and Development in
  Information Retrieval (SIGIR)}, pp.\  173--176, 2019.

\bibitem[Rusak et~al.(2020)Rusak, Schott, Zimmermann, Bitterwolf, Bringmann,
  Bethge, and Brendel]{rusak2020simple}
Rusak, E., Schott, L., Zimmermann, R.~S., Bitterwolf, J., Bringmann, O.,
  Bethge, M., and Brendel, W.
\newblock A simple way to make neural networks robust against diverse image
  corruptions.
\newblock In \emph{European Conference on Computer Vision (ECCV)}, pp.\
  53--69. Springer, 2020.

\bibitem[Tian et~al.(2020)Tian, Sun, Poole, Krishnan, Schmid, and
  Isola]{tian2020makes}
Tian, Y., Sun, C., Poole, B., Krishnan, D., Schmid, C., and Isola, P.
\newblock What makes for good views for contrastive learning?
\newblock In \emph{International Conference on Neural Information Processing
  Systems (NeurIPS)}, pp.\  6827--6839, 2020.

\bibitem[Veli{\v{c}}kovi{\'c} et~al.(2018)Veli{\v{c}}kovi{\'c}, Cucurull,
  Casanova, Romero, Lio, and Bengio]{velivckovic2017graph}
Veli{\v{c}}kovi{\'c}, P., Cucurull, G., Casanova, A., Romero, A., Lio, P., and
  Bengio, Y.
\newblock Graph attention networks.
\newblock In \emph{International Conference on Learning Representations
  (ICLR)}, 2018.

\bibitem[Wang et~al.(2022)Wang, Lin, Yang, Sun, Yue, and
  Shahabi]{wang2022hagen}
Wang, C., Lin, Z., Yang, X., Sun, J., Yue, M., and Shahabi, C.
\newblock Hagen: Homophily-aware graph convolutional recurrent network for
  crime forecasting.
\newblock In \emph{International Conference on Artificial Intelligence (AAAI)},
  volume~36, pp.\  4193--4200, 2022.

\bibitem[Wang \& Li(2017)Wang and Li]{wang2017region}
Wang, H. and Li, Z.
\newblock Region representation learning via mobility flow.
\newblock In \emph{Conference on Information and Knowledge Management (CIKM)},
  pp.\  237--246, 2017.

\bibitem[Wang et~al.(2019)Wang, Zhang, Liu, Chen, Xu, Fardad, and
  Li]{wang2019towards}
Wang, J., Zhang, T., Liu, S., Chen, P.-Y., Xu, J., Fardad, M., and Li, B.
\newblock Towards a unified min-max framework for adversarial exploration and
  robustness.
\newblock In \emph{Conference on Neural Information Processing Systems
  (NeurIPS)}, 2019.

\bibitem[Wu et~al.(2022)Wu, Yan, Fan, Pan, Zhu, Zheng, Cheng, and
  Wang]{wu2022multi_graph}
Wu, S., Yan, X., Fan, X., Pan, S., Zhu, S., Zheng, C., Cheng, M., and Wang, C.
\newblock Multi-graph fusion networks for urban region embedding.
\newblock In \emph{International Joint Conference on Artificial Intelligence
  (IJCAI)}, 2022.

\bibitem[Xia et~al.()Xia, Huang, Xu, Dai, Bo, Zhang, and Chen]{xia21spatial}
Xia, L., Huang, C., Xu, Y., Dai, P., Bo, L., Zhang, X., and Chen, T.
\newblock Spatial-temporal sequential hypergraph network for crime prediction
  with dynamic multiplex relation learning.
\newblock In \emph{International Joint Conference on Artificial Intelligence
  (IJCAI)}, pp.\  1631--1637.

\bibitem[Xia et~al.(2023)Xia, Huang, Shi, and Xu]{xia2023graph}
Xia, L., Huang, C., Shi, J., and Xu, Y.
\newblock Graph-less collaborative filtering.
\newblock In \emph{The Web Conference (WWW)}, pp.\  17--27, 2023.

\bibitem[Xu et~al.(2021)Xu, Cheng, Luo, Chen, and Zhang]{xu2021infogcl}
Xu, D., Cheng, W., Luo, D., Chen, H., and Zhang, X.
\newblock Infogcl: Information-aware graph contrastive learning.
\newblock \emph{International Conference on Neural Information Processing
  Systems (NeurIPS)}, 34:\penalty0 30414--30425, 2021.

\bibitem[Xu et~al.(2019)Xu, Chen, Liu, Chen, Weng, Hong, and
  Lin]{xu2019topology}
Xu, K., Chen, H., Liu, S., Chen, P.-Y., Weng, T.-W., Hong, M., and Lin, X.
\newblock Topology attack and defense for graph neural networks: An
  optimization perspective.
\newblock \emph{arXiv preprint arXiv:1906.04214}, 2019.

\bibitem[Yang et~al.(2022)Yang, Liu, and Zhao]{yang2022getnext}
Yang, S., Liu, J., and Zhao, K.
\newblock Getnext: trajectory flow map enhanced transformer for next poi
  recommendation.
\newblock In \emph{International Conference on Research and Development in
  Information Retrieval (SIGIR)}, pp.\  1144--1153, 2022.

\bibitem[Yao et~al.(2018)Yao, Fu, Liu, Hu, and Xiong]{yao2018representing}
Yao, Z., Fu, Y., Liu, B., Hu, W., and Xiong, H.
\newblock Representing urban functions through zone embedding with human
  mobility patterns.
\newblock In \emph{International Joint Conference on Artificial Intelligence
  (IJCAI)}, 2018.

\bibitem[Yi et~al.(2016)Yi, Zheng, Zhang, and Li]{yi2016st}
Yi, X., Zheng, Y., Zhang, J., and Li, T.
\newblock St-mvl: filling missing values in geo-sensory time series data.
\newblock In \emph{International Joint Conference on Artificial Intelligence
  (IJCAI)}, 2016.

\bibitem[Yin et~al.(2022)Yin, Wang, Huang, Xiong, and Zhang]{yin2022autogcl}
Yin, Y., Wang, Q., Huang, S., Xiong, H., and Zhang, X.
\newblock Autogcl: Automated graph contrastive learning via learnable view
  generators.
\newblock In \emph{International Conference on Artificial Intelligence (AAAI)},
  volume~36, pp.\  8892--8900, 2022.

\bibitem[You et~al.(2020)You, Chen, Sui, Chen, Wang, and Shen]{you2020graph}
You, Y., Chen, T., Sui, Y., Chen, T., Wang, Z., and Shen, Y.
\newblock Graph contrastive learning with augmentations.
\newblock \emph{International Conference on Neural Information Processing
  Systems (NeurIPS)}, 33:\penalty0 5812--5823, 2020.

\bibitem[Yu et~al.(2018)Yu, Yin, and Zhu]{yu2017spatio}
Yu, B., Yin, H., and Zhu, Z.
\newblock Spatio-temporal graph convolutional networks: A deep learning
  framework for traffic forecasting.
\newblock In \emph{International Joint Conference on Artificial Intelligence
  (IJCAI)}, 2018.

\bibitem[Zhang et~al.(2021{\natexlab{a}})Zhang, Li, Li, and
  Hui]{zhang2021multi}
Zhang, M., Li, T., Li, Y., and Hui, P.
\newblock Multi-view joint graph representation learning for urban region
  embedding.
\newblock In \emph{International Joint Conference on Artificial Intelligence
  (IJCAI)}, pp.\  4431--4437, 2021{\natexlab{a}}.

\bibitem[Zhang et~al.(2020)Zhang, Chang, Meng, Xiang, and Pan]{zhang2020spatio}
Zhang, Q., Chang, J., Meng, G., Xiang, S., and Pan, C.
\newblock Spatio-temporal graph structure learning for traffic forecasting.
\newblock In \emph{International Conference on Artificial Intelligence (AAAI)},
  volume~34, pp.\  1177--1185, 2020.

\bibitem[Zhang et~al.(2023)Zhang, Huang, Xia, Wang, Li, and
  Yiu]{zhang2023automated}
Zhang, Q., Huang, C., Xia, L., Wang, Z., Li, Z., and Yiu, S.
\newblock Automated spatio-temporal graph contrastive learning.
\newblock In \emph{Proceedings of the ACM Web Conference 2023}, pp.\  295--305,
  2023.

\bibitem[Zhang et~al.(2021{\natexlab{b}})Zhang, Huang, Xu, Xia, Dai, Bo, Zhang,
  and Zheng]{zhang2021traffic}
Zhang, X., Huang, C., Xu, Y., Xia, L., Dai, P., Bo, L., Zhang, J., and Zheng,
  Y.
\newblock Traffic flow forecasting with spatial-temporal graph diffusion
  network.
\newblock In \emph{AAAI Conference on Artificial Intelligence (AAAI)}, pp.\
  15008--15015, 2021{\natexlab{b}}.

\bibitem[Zhang et~al.(2019)Zhang, Fu, Wang, Li, and Zheng]{zhang2019unifying}
Zhang, Y., Fu, Y., Wang, P., Li, X., and Zheng, Y.
\newblock Unifying inter-region autocorrelation and intra-region structures for
  spatial embedding via collective adversarial learning.
\newblock In \emph{International Conference on Knowledge Discovery and Data
  Mining (KDD)}, pp.\  1700--1708, 2019.

\bibitem[Zheng et~al.(2020)Zheng, Fan, Wang, and Qi]{zheng2020gman}
Zheng, C., Fan, X., Wang, C., and Qi, J.
\newblock Gman: A graph multi-attention network for traffic prediction.
\newblock In \emph{International Conference on Artificial Intelligence (AAAI)},
  volume~34, pp.\  1234--1241, 2020.

\bibitem[Zhou et~al.(2019)Zhou, Mascolo, and Zhao]{zhou2019topic}
Zhou, X., Mascolo, C., and Zhao, Z.
\newblock Topic-enhanced memory networks for personalised point-of-interest
  recommendation.
\newblock In \emph{Conference on Knowledge Discovery and Data Mining (KDD)},
  pp.\  3018--3028, 2019.

\bibitem[Zhou et~al.(2020)Zhou, Wang, Xie, Chen, and Liu]{zhou2020riskoracle}
Zhou, Z., Wang, Y., Xie, X., Chen, L., and Liu, H.
\newblock Riskoracle: a minute-level citywide traffic accident forecasting
  framework.
\newblock In \emph{AAAI Conference on Artificial Intelligence (AAAI)},
  volume~34, pp.\  1258--1265, 2020.

\bibitem[Zhou et~al.(2023)Zhou, Lin, Yang, BAI, Wang, et~al.]{zhou2023greto}
Zhou, Z., Lin, G., Yang, K., BAI, L., Wang, Y., et~al.
\newblock Greto: Remedying dynamic graph topology-task discordance via target
  homophily.
\newblock In \emph{International Conference on Learning Representations
  (ICLR)}, 2023.

\bibitem[Zhu et~al.(2021)Zhu, Xu, Yu, Liu, Wu, and Wang]{zhu2021graph}
Zhu, Y., Xu, Y., Yu, F., Liu, Q., Wu, S., and Wang, L.
\newblock Graph contrastive learning with adaptive augmentation.
\newblock In \emph{The Web Conference (WWW)}, pp.\  2069--2080, 2021.

\end{thebibliography}
\bibliographystyle{icml2023}

\newpage
\appendix
\onecolumn
\section{Appendix}
\label{sec:appendix}

\subsection{Description of Baselines}
\label{sec:baseline}


To provide a comprehensive evaluation of our \model\ model, we compare it with many baseline methods from three different research lines. The first research line includes graph representation approaches, which are commonly used for embedding graph-structured data. The second research line includes graph contrastive learning models, which have shown promising results in learning discriminative representations for graphs. The third research line includes spatial-temporal region representation methods, which are specifically designed for embedding spatial-temporal data.

\noindent \underline{\bf Network Embedding/GNN Approaches}. To evaluate the effectiveness of our \model\ model, we compare it with several representative network embedding and graph neural network models. We apply these models on our region graph $\mathcal{G}$ to generate region embeddings. The details of each baseline is described as follows: \textbf{Node2vec}~\cite{grover2016node2vec}: It encodes graph structural information via random walk-based Skip-gram. \textbf{GCN}~\cite{kipf2016semi}: It performs the convolution-based message passing between neighbor nodes along the edges for embedding refinement. \textbf{GraphSage}~\cite{hamilton2017inductive}: It is a graph neural architecture that enables information aggregation from the sampled sub-graph structures. \textbf{GAE}~\cite{kipf2016variational}: Graph Auto-encoder maps nodes into a latent embedding space with the input reconstruction objective over the graph structures. \textbf{GAT}~\cite{velivckovic2017graph}: Graph Attention Network enhances the discrimination ability of GNNs by differentiating the relevance degrees among neighboring nodes. 



\noindent \underline{\bf Graph Contrastive Learning Methods.} In addition to the above-mentioned graph representation and GNN-based models, we also compare our \model\ model with two graph contrastive learning models, namely, \textbf{GraphCL}~\cite{you2020graph}: This method creates multiple contrastive views for augmentation based on mutual information maximization. The embedding consistency is aimed to be achieved among different correlated views. \textbf{RGCL}~\cite{li2022let}: This is a state-of-the-art graph contrastive learning approach that performs data augmentation based on the designed rationale generator.


\noindent \underline{\bf Spatial-Temporal Region Representation Models.} we also compare it with state-of-the-art spatial-temporal representation methods for region embedding. These methods are as follows:
\textbf{POI}: It utilizes POI attributes to represent spatial regions via TF-IDF tokenization given POI matrix.
\textbf{HDGE}~\cite{wang2017region}: It uses human trajectories to generate a crowd flow graph and embeds regions into latent vectors to preserve graph structural information.
\textbf{ZE-Mob}~\cite{yao2018representing}: In this method, region correlations are captured with the consideration of human mobility and taxi moving traces for producing embeddings.
\textbf{MV-PN}~\cite{fu2019efficient}: It is an encoder-decoder network to model intra-region and inter-region correlations.
\textbf{CGAL}~\cite{zhang2019unifying}: A graph-regularized adversarial learning method that considers graph-structured pairwise relations to embed regions into latent space.
\textbf{MVURE}~\cite{zhang2021multi}: It leverages the graph attention mechanism to model region correlations with inherent region attributes and human mobility data.
\textbf{MGFN}~\cite{wu2022multi_graph}: It encodes region embeddings with multi-level cross-attention to aggregate information for both intra-pattern and inter-pattern. \vspace{-0.2in} 


\begin{table*}[h]
\center
\setlength{\abovecaptionskip}{0cm}
\setlength{\belowcaptionskip}{0cm}
\setlength{\tabcolsep}{1.5pt}
\small
\caption{Overall performance comparison in crime prediction on both Chicago and NYC datasets.}
\label{fig:crime_all}
\begin{tabular}{|c|cccccccc|cccccccc|}
\hline
           & \multicolumn{8}{c|}{Chicago}                                                                                                                                                                         & \multicolumn{8}{c|}{New York City}                                                                                                                                                                   \\ \cline{2-17}
Model      & \multicolumn{2}{c|}{Theft}                           & \multicolumn{2}{c|}{Battery}                         & \multicolumn{2}{c|}{Assault}                         & \multicolumn{2}{c|}{Damage}     & \multicolumn{2}{c|}{Burglary}                        & \multicolumn{2}{c|}{Larceny}                         & \multicolumn{2}{c|}{Robbery}                         & \multicolumn{2}{c|}{Assault}    \\ \cline{2-17}
           & \multicolumn{1}{c|}{MAE} & \multicolumn{1}{c|}{MPAE} & \multicolumn{1}{c|}{MAE} & \multicolumn{1}{c|}{MPAE} & \multicolumn{1}{c|}{MAE} & \multicolumn{1}{c|}{MPAE} & \multicolumn{1}{c|}{MAE} & MPAE & \multicolumn{1}{c|}{MAE} & \multicolumn{1}{c|}{MPAE} & \multicolumn{1}{c|}{MAE} & \multicolumn{1}{c|}{MPAE} & \multicolumn{1}{c|}{MAE} & \multicolumn{1}{c|}{MPAE} & \multicolumn{1}{c|}{MAE} & MPAE \\ \hline \hline
Node2vec   & \multicolumn{1}{c|}{1.1378}    & \multicolumn{1}{c|}{0.9862}     & \multicolumn{1}{c|}{1.7655}    & \multicolumn{1}{c|}{0.8970}     & \multicolumn{1}{c|}{1.9631}    & \multicolumn{1}{c|}{0.9714}     & \multicolumn{1}{c|}{1.9015}    &0.9657      & \multicolumn{1}{c|}{4.9447}    & \multicolumn{1}{c|}{0.8092}     & \multicolumn{1}{c|}{0.7272}    & \multicolumn{1}{c|}{0.6532}     & \multicolumn{1}{c|}{1.0566}    & \multicolumn{1}{c|}{0.8040}     & \multicolumn{1}{c|}{1.2411}    &0.9967      \\ \hline
GCN      & \multicolumn{1}{c|}{1.1065}    & \multicolumn{1}{c|}{0.9643}     & \multicolumn{1}{c|}{1.3012}    & \multicolumn{1}{c|}{0.8094}     & \multicolumn{1}{c|}{1.5431}    & \multicolumn{1}{c|}{0.8094}     & \multicolumn{1}{c|}{1.5031}    &0.8056      & \multicolumn{1}{c|}{4.6993}    & \multicolumn{1}{c|}{0.7912}     & \multicolumn{1}{c|}{0.49994}    & \multicolumn{1}{c|}{0.4178}     & \multicolumn{1}{c|}{1.0655}    & \multicolumn{1}{c|}{0.8004}     & \multicolumn{1}{c|}{1.2407}    &0.9890     \\ \hline
GAT        & \multicolumn{1}{c|}{1.1123}    & \multicolumn{1}{c|}{0.9759}     & \multicolumn{1}{c|}{1.3215}    & \multicolumn{1}{c|}{0.8344}     & \multicolumn{1}{c|}{1.5892}    & \multicolumn{1}{c|}{0.8241}     & \multicolumn{1}{c|}{1.5387}    &0.8277      & \multicolumn{1}{c|}{4.7055}    & \multicolumn{1}{c|}{0.7944}     & \multicolumn{1}{c|}{0.5023}    & \multicolumn{1}{c|}{0.4019}     & \multicolumn{1}{c|}{1.0653}    & \multicolumn{1}{c|}{0.8027}     & \multicolumn{1}{c|}{1.2403}    &0.9949      \\ \hline
GraphSage        & \multicolumn{1}{c|}{1.1231}    & \multicolumn{1}{c|}{0.9790}     & \multicolumn{1}{c|}{1.3574}    & \multicolumn{1}{c|}{0.8561}     & \multicolumn{1}{c|}{1.6016}    & \multicolumn{1}{c|}{0.8563}     & \multicolumn{1}{c|}{1.5761}    &0.8432      & \multicolumn{1}{c|}{4.7313}    & \multicolumn{1}{c|}{0.8066}     & \multicolumn{1}{c|}{0.5213}    & \multicolumn{1}{c|}{0.4314}     & \multicolumn{1}{c|}{1.0719}    & \multicolumn{1}{c|}{0.8110}     & \multicolumn{1}{c|}{1.2418}    &0.9965      \\ \hline
GAE        & \multicolumn{1}{c|}{1.1043}    & \multicolumn{1}{c|}{0.9614}     & \multicolumn{1}{c|}{1.3065}    & \multicolumn{1}{c|}{0.7984}     & \multicolumn{1}{c|}{1.5379}    & \multicolumn{1}{c|}{0.7914}     & \multicolumn{1}{c|}{1.4986}    &0.8033      & \multicolumn{1}{c|}{4.7013}    & \multicolumn{1}{c|}{0.7910}     & \multicolumn{1}{c|}{0.5012}    & \multicolumn{1}{c|}{0.4289}     & \multicolumn{1}{c|}{1.0679}    & \multicolumn{1}{c|}{0.8012}     & \multicolumn{1}{c|}{1.2405}    &0.9958      \\ \hline
GraphCL  & \multicolumn{1}{c|}{0.9994}    & \multicolumn{1}{c|}{0.8992}     & \multicolumn{1}{c|}{1.0578}    & \multicolumn{1}{c|}{0.6174}     & \multicolumn{1}{c|}{1.2938}    & \multicolumn{1}{c|}{0.5683}     & \multicolumn{1}{c|}{1.2457}    &0.6181      & \multicolumn{1}{c|}{4.5047}    & \multicolumn{1}{c|}{0.7288}     & \multicolumn{1}{c|}{0.3834}    & \multicolumn{1}{c|}{0.3091}     & \multicolumn{1}{c|}{1.0498}    & \multicolumn{1}{c|}{0.5961}     & \multicolumn{1}{c|}{0.7724}    &0.6929      \\ \hline
RGCL  & \multicolumn{1}{c|}{0.9377}   & \multicolumn{1}{c|}{0.8779}     & \multicolumn{1}{c|}{1.0356}    & \multicolumn{1}{c|}{0.5865}     & \multicolumn{1}{c|}{1.2764}    & \multicolumn{1}{c|}{0.5457}     & \multicolumn{1}{c|}{1.2336}    &0.5975      & \multicolumn{1}{c|}{4.4712}    & \multicolumn{1}{c|}{0.7064}     & \multicolumn{1}{c|}{0.3652}    & \multicolumn{1}{c|}{0.2832}     & \multicolumn{1}{c|}{0.9987}    & \multicolumn{1}{c|}{0.5769}     & \multicolumn{1}{c|}{0.7512}    &0.6743      \\ \hline
POI        & \multicolumn{1}{c|}{0.9733}    & \multicolumn{1}{c|}{0.9341}     & \multicolumn{1}{c|}{1.1065}    & \multicolumn{1}{c|}{0.7513}     & \multicolumn{1}{c|}{1.4089}    & \multicolumn{1}{c|}{0.7541}     & \multicolumn{1}{c|}{1.4076}    &0.7697      & \multicolumn{1}{c|}{4.6939}    & \multicolumn{1}{c|}{0.7825}     & \multicolumn{1}{c|}{0.4969}    & \multicolumn{1}{c|}{0.4172}     & \multicolumn{1}{c|}{1.0660}    & \multicolumn{1}{c|}{0.7970}     & \multicolumn{1}{c|}{1.2400}    &0.9943     \\ \hline
HDGE       & \multicolumn{1}{c|}{0.9545}    & \multicolumn{1}{c|}{0.9012}     & \multicolumn{1}{c|}{1.0887}    & \multicolumn{1}{c|}{0.7389}     & \multicolumn{1}{c|}{1.3970}    & \multicolumn{1}{c|}{0.7217}     & \multicolumn{1}{c|}{1.3768}    &0.7349      & \multicolumn{1}{c|}{4.5658}    & \multicolumn{1}{c|}{0.7160}     & \multicolumn{1}{c|}{0.4734}    & \multicolumn{1}{c|}{0.3930}     & \multicolumn{1}{c|}{1.0507}    & \multicolumn{1}{c|}{0.6731}     & \multicolumn{1}{c|}{1.1551}    &0.9870     \\ \hline
ZE-Mob      & \multicolumn{1}{c|}{1.0983}    & \multicolumn{1}{c|}{0.9547}     & \multicolumn{1}{c|}{1.3142}    & \multicolumn{1}{c|}{0.8236}     & \multicolumn{1}{c|}{1.5345}    & \multicolumn{1}{c|}{0.8163}     & \multicolumn{1}{c|}{1.5138}    &0.8273      & \multicolumn{1}{c|}{4.7570}    & \multicolumn{1}{c|}{0.8013}     & \multicolumn{1}{c|}{0.5186}    & \multicolumn{1}{c|}{0.4307}     & \multicolumn{1}{c|}{1.0722}    & \multicolumn{1}{c|}{0.8093}     & \multicolumn{1}{c|}{1.1403}    &0.9975      \\ \hline
MV-PN     & \multicolumn{1}{c|}{0.9613}    & \multicolumn{1}{c|}{0.9146}     & \multicolumn{1}{c|}{1.0946}    & \multicolumn{1}{c|}{0.7452}     & \multicolumn{1}{c|}{1.4013}    & \multicolumn{1}{c|}{0.7393}     & \multicolumn{1}{c|}{1.3582}    &0.7218      & \multicolumn{1}{c|}{4.6329}    & \multicolumn{1}{c|}{0.7502}     & \multicolumn{1}{c|}{0.4213}    & \multicolumn{1}{c|}{0.3708}     & \multicolumn{1}{c|}{1.0642}    & \multicolumn{1}{c|}{0.7840}     & \multicolumn{1}{c|}{1.1091}    &0.9982      \\ \hline
CGAL      & \multicolumn{1}{c|}{0.9589}    & \multicolumn{1}{c|}{0.9014}     & \multicolumn{1}{c|}{1.0897}    & \multicolumn{1}{c|}{0.7403}     & \multicolumn{1}{c|}{1.3995}    & \multicolumn{1}{c|}{0.7345}     & \multicolumn{1}{c|}{1.3698}    &0.7296      & \multicolumn{1}{c|}{4.6013}    & \multicolumn{1}{c|}{0.7203}     & \multicolumn{1}{c|}{0.4113}    & \multicolumn{1}{c|}{0.3651}     & \multicolumn{1}{c|}{1.0714}    & \multicolumn{1}{c|}{0.7765}     & \multicolumn{1}{c|}{1.1009}    &0.9894      \\ \hline
MVURE      & \multicolumn{1}{c|}{0.9365}    & \multicolumn{1}{c|}{0.8910}     & \multicolumn{1}{c|}{1.0631}    & \multicolumn{1}{c|}{0.6957}     & \multicolumn{1}{c|}{1.3709}    & \multicolumn{1}{c|}{0.6375}     & \multicolumn{1}{c|}{1.3037}    &0.6567      & \multicolumn{1}{c|}{4.5907}    & \multicolumn{1}{c|}{0.7144}     & \multicolumn{1}{c|}{0.4077}    & \multicolumn{1}{c|}{0.3262}     & \multicolumn{1}{c|}{1.0578}    & \multicolumn{1}{c|}{0.5889}     & \multicolumn{1}{c|}{0.8410}    &0.6943     \\ \hline
MGFN      & \multicolumn{1}{c|}{0.9231}    & \multicolumn{1}{c|}{0.9015}     & \multicolumn{1}{c|}{1.0804}    & \multicolumn{1}{c|}{0.5824}     & \multicolumn{1}{c|}{1.3016}    & \multicolumn{1}{c|}{0.6072}     & \multicolumn{1}{c|}{1.2563}    &0.6503      & \multicolumn{1}{c|}{4.5646}    & \multicolumn{1}{c|}{0.7994}     & \multicolumn{1}{c|}{0.4285}    & \multicolumn{1}{c|}{0.3084}     & \multicolumn{1}{c|}{1.0475}    & \multicolumn{1}{c|}{0.6310}     & \multicolumn{1}{c|}{0.8319}    &0.7096     \\ \hline \hline 
\model & \multicolumn{1}{c|}{\textbf{0.9107}}    & \multicolumn{1}{c|}{\textbf{0.8424}}     & \multicolumn{1}{c|}{\textbf{0.9969}}    & \multicolumn{1}{c|}{\textbf{0.5618}}     & \multicolumn{1}{c|}{\textbf{1.2068}}    & \multicolumn{1}{c|}{\textbf{0.4944}}     & \multicolumn{1}{c|}{\textbf{1.1056}}    &\textbf{0.5438} & \multicolumn{1}{c|}{\textbf{4.3095}}    & \multicolumn{1}{c|}{\textbf{0.6504}}     & \multicolumn{1}{c|}{\textbf{0.2958}}    & \multicolumn{1}{c|}{\textbf{0.1924}}     & \multicolumn{1}{c|}{\textbf{0.8809}}    & \multicolumn{1}{c|}{\textbf{0.3902}}     & \multicolumn{1}{c|}{\textbf{0.7501}}    &\textbf{0.6510}
\\ \hline
\end{tabular}
\vspace{-1.0em}
\end{table*}

\subsection{Learning Process of \model}


As shown in Algorithm~\ref{alg:learn_alg}, our \model\ model first constructs the multi-view region-wise graph using the three data views (POI, mobility, and distance). Then it learns urban region embeddings with the space-time message passing paradigm. The variational graph autoencoder is employed for data augmentations, based on which \model\ conducts adversarial learning against hard samples, cross-view contrastive learning, and information regulation.

\begin{algorithm}[h]
	\caption{Learning Process of \model}
	\label{alg:learn_alg}
	\LinesNumbered
	\KwIn{Region Point-of-Interest matrix $\mathcal{P}$, mobility trajectories $\mathcal{M}$, region geographical positions $\mathcal{D}$, learning rate $\eta$, training epochs $E$.}
	\KwOut{Regional embeddings $\tilde{\textbf{H}}$}
	Initialize model parameters in each module\;
 
	Initialize POI-based region embeddings through $\bar{\textbf{E}} = \text{MLP}(\text{Skip-gram}(\mathcal{P}))$\;
 
	Capture region-wise correlations with self-attention  to acquire the initial embeddings $\textbf{E}$
 
	Construct the multi-view region graph $\mathcal{G}$
 
    \For{$e=1$ to $E$}{
        Conduct spatial-temporal message passing to get low-dimensional region embeddings $\textbf{H}$;
        
        Apply variational graph augmentation in the latent space twice for $\tilde{\textbf{H}}$ and $\tilde{\textbf{H}}'$;
        
        Apply minimax adversarial learning in the latent space $\tilde{\textbf{H}}$ and obtain the hard sample $\textbf{H}_{\text{adv}}$ and conduct adversarial contrastive learning;
        
        Conduct cross-view contrastive learning between pairs of data views $\tilde{\textbf{H}}_{p}$ and $\tilde{\textbf{H}}_{m}$, $\tilde{\textbf{H}}_{p}$ and $\tilde{\textbf{H}}_{s}$, and $\tilde{\textbf{H}}_{m}$ and $\tilde{\textbf{H}}_{s}$ following Eq 8 and Eq 9;
        
        Apply contrative learning between $\tilde{\textbf{H}}$ and $\tilde{\textbf{H}'}$;
        Apply on inforamtion regularation on the results of contrastive learning following Eq 11;
        
        Calculate the loss of \model\ folloing the Eq 12;
    }
    \For{each parameter ${\theta}$ in $\mathbf{\Theta}$}{
        ${\theta} = {\theta} - \eta\cdot \frac{\partial \mathcal{L}}{\partial{\theta}}$;
        
    }
    \Return all parameters $\mathbf{\Theta}$
\end{algorithm}

\subsection{In-Depth Analysis of \model}
\textbf{Projection in Adversarial Contrasting}. 
In this section, we aim to provide further discussion on the projection operation $\prod \limits_{\mathcal{K}_{\hat{\mathbf{A}}'}}(\mathbf{Y})$ in projected gradient descent (PGD) of our adversarial contrasting learning. In general, our adversarial contrasting module is designed to enable our \model\ to be robust by distilling hard samples for improving model optimization. Inspired by the works in~\cite{xu2019topology,feng2022adversarial}, the projection operation generates similar-form solutions as projected gradient descent. We first transform the optimization objective from the PGD attack to the projection operation as:
\begin{align}
\label{eq:oper}
&\mathop{\text{minimize}}\limits_{\mathcal{K}_{\hat{\mathbf{A}}'}}~~~~ \frac{1}{2} ||\mathcal{K}_{\hat{\mathbf{A}}'} - \textbf{Y}||^2_{2} + \mathcal{N}_{[0,1]}(\mathcal{K}_{\hat{\mathbf{A}}'}) \nonumber\\
&\text{subject to}~~~~\textbf{1}^{\mathrm{T}}\mathcal{K}_{\hat{\mathbf{A}}'} \leq \triangle_{\hat{\mathbf{A}}'}
\end{align}
where when $\mathcal{K}_{\hat{\mathbf{A}}'} \in [0,1]^n$, $\mathcal{N}_{[0,1]}(\mathcal{K}_{\hat{\mathbf{A}}'}) = 0$, otherwise 
$\mathcal{N}_{[0,1]}(\mathcal{K}_{\hat{\mathbf{A}}'}) = \infty$. Thus, the Lagrangian function of Eq~\ref{eq:oper} is expressed as follows~\cite{boyd2004convex}:
\begin{align}
\label{eq:oper_lagr}
\mathop{\text{minimize}}\limits_{\mathcal{K}_{\hat{\mathbf{A}}'}}~~~~ \frac{1}{2} ||\mathcal{K}_{\hat{\mathbf{A}}'} - \textbf{Y}||^2_{2} + \mathcal{N}_{[0,1]}(\mathcal{K}_{\hat{\mathbf{A}}'}) + \vartheta(\textbf{1}^{\mathrm{T}}\mathcal{K}_{\hat{\mathbf{A}}'}-\triangle_{\hat{\mathbf{A}}'})
\end{align}
where $\vartheta$ is the dual variable and $\vartheta\geq 0$. Thus the minimization of Eq~\ref{eq:oper_lagr} is $\mathcal{K}_{\hat{\mathbf{A}}'} = P_{[0,1]}(\textbf{Y}- \vartheta\textbf{1})$, which provides solutions for PGD attack problem in our adversarial contrastive training process.

\noindent \textbf{Robust Adversarial Training with PGD Attack}. In this part, we aim to clarify the rationale of adopting PGD Attack as the solution for the adversarial training ($\mathcal{L}_{\text{adv}}$ in Eq.~\ref{eq:final_loss}) of spatial-temporal graph neural networks, so as to achieve model robustness. To this end, we first define our PGD attack problem for our spatial-temporal graph neural network as follows:
\begin{align}
\label{eq:attack_generation}
\mathop{\text{minimize}}\limits_{\mathbf{k} \in \mathcal{K}}\mathop{\text{maximize}}\limits_{\mathbf{W}} - g(\mathbf{k}, \mathbf{W})
\end{align}
\noindent where Eq.~\ref{eq:attack_generation} is the min-max form of the PGD attack problem, where $g$ is a cross-entropy-based loss function that measures the discrepancy between the model prediction and the true label. Here, we present how to approximate the PGD attack problem as our target adversarial training objective. By doing so, the PGD attack can serve as an effective solution for our adversarial augmentation with hard sample mining over our generated multi-view spatial-temporal graph $\mathcal{G}$.



\noindent\textbf{Proof:}
Given $\mathcal{K}$ and $\mathbf{W}$, we need to optimize the following formula in adversarial training:
\begin{align}
\label{eq:opti_1}
\mathop{\text{minimize}}\limits_{\mathbf{W}}\mathop{\text{maximize}}\limits_{\mathbf{k} \in \mathcal{K}} - g(\mathbf{k}, \mathbf{W})
\end{align}
where $\mathbf{W}$ is the weight matrix of the spatial-temporal graph neural network. Given a general loss function $g$ of the attack operation on the spatial-temporal neural networks, Eq.~\ref{eq:opti_1} is equivalent to $\mathop{\text{maximize}}\limits_{\mathbf{W}}\mathop{\text{minimize}}\limits_{\mathbf{k} \in \mathcal{K}} g(\mathbf{k}, \mathbf{W})$, which can be proved as follows. We introduce an epigraph variable $h$~\cite{boyd2004convex} to rewrite Eq.~\ref{eq:opti_1} as:
\begin{align}
\label{eq:opti_2}
&\mathop{\text{minimize}}\limits_{\mathbf{W}, h}~~~~ h \nonumber\\
&\text{subject to}~~~~ -g(\mathbf{k}, \mathbf{W}) \leq h, \forall \mathbf{k} \in \mathcal{K}
\end{align}
We set $d := -h$, Eq~\ref{eq:opti_2} is expressed as:
\begin{align}
\label{eq:opti_3}
&\mathop{\text{maximize}}\limits_{\mathbf{W}, d}~~~~ d \nonumber\\
&\text{subject to}~~~~ g(\mathbf{k}, \mathbf{W}) \geq d, \forall \mathbf{k} \in \mathcal{K}
\end{align}
Then removing the epigraph variable $d$, Eq~\ref{eq:opti_3} is equal to 
$\mathop{\text{maximize}}\limits_{\mathbf{W}}\mathop{\text{minimize}}\limits_{\mathbf{k} \in \mathcal{K}} g(\mathbf{k}, \mathbf{W})$. According max-min inequality~\cite{boyd2004convex}, we can proof that:
\begin{align}
\label{eq:final}
\mathop{\text{maximize}}\limits_{\mathbf{W}}\mathop{\text{minimize}}\limits_{\mathbf{k} \in \mathcal{K}} g(\mathbf{k}, \mathbf{W}) \leq \mathop{\text{minimize}}\limits_{\mathbf{k} \in \mathcal{K}}\mathop{\text{maximize}}\limits_{\mathbf{W}} - g(\mathbf{k}, \mathbf{W}) = \mathcal{L}_{\text{adv}}
\end{align}
Based on the discussion above, it can be inferred that the PGD attack generation problem is a suitable approximation for adversarial training when using the min-max optimization paradigm. This approach, coupled with projection operations, can enhance spatial-temporal adversarial contrastive learning, allowing for the identification of hard samples for self-supervision.



\subsection{Category-Specific Crime Prediction Results}
In the supplementary materials, we present detailed evaluation results for different crime types in Chicago and New York City. We evaluate different methods in terms of MAE and MAPE, with ST-SHN as the backbone method for all methods. Our \model\ framework consistently achieves the best results across all crime categories for both cities, as shown in Table~\ref{fig:crime_all}. These results demonstrate the significant benefits of our spatial-temporal graph learning framework. We attribute the superiority of our approach to the factors, including the graph encoding on the multi-view region graph, which effectively extracts useful regional features for region representation. Additionally, the various contrastive learning tasks, including autoencoder-based adaptive contrastive learning, adversarial contrastive learning with hard negative generation, and cross-view contrastive learning, further enhancing the performance of our approach.



\end{document}